\def\year{2020}\relax
\documentclass[letterpaper]{article} 
\usepackage{cite}
\usepackage{aaai20}  
\usepackage{times}  
\usepackage{helvet} 
\usepackage{courier}  
\usepackage[hyphens]{url}  
\usepackage{graphicx} 
\usepackage{graphicx}  
\usepackage{subfig}
\usepackage{epstopdf}			
\usepackage[mathscr]{eucal}         
\usepackage{amsbsy}                 
\usepackage{algorithm}
\usepackage{algorithmic}
\usepackage{amsmath,amssymb,amsthm,float}
\usepackage{enumitem}
\usepackage{mdwmath}
\usepackage{mdwtab}
\usepackage{enumitem}

\numberwithin{theorem}{section}

\newcommand{\bm}{\boldsymbol}

\newcommand{\ten}[1]{ \boldsymbol{\mathcal #1}}
\newcommand{\bbR}[1]{\mathbb{R}^{#1}}

\usepackage{booktabs}

\newenvironment{mytabular2}{\bgroup\scriptsize  \tabular}{\endtabular\egroup}

\title{Block Hankel Tensor ARIMA for Multiple Short Time Series Forecasting}
\author{
	Qiquan Shi,\textsuperscript{\rm 1}
	 Jiaming Yin,\textsuperscript{\rm 2}${}^{\ *}$
Jiajun  Cai,\textsuperscript{\rm 3}~\thanks{Work done during an internship at Huawei.} 
	Andrzej Cichocki,\textsuperscript{\rm 4}
Tatsuya Yokota,\textsuperscript{\rm 5}\\\Large\textbf{
Lei Chen,\textsuperscript{\rm 1}
Mingxuan Yuan,\textsuperscript{\rm 1}
Jia Zeng\textsuperscript{\rm 1}}
\\
	\textsuperscript{\rm 1}Huawei Noah's Ark Lab, Hong Kong, China,
	\textsuperscript{\rm 2}Tongji University, Shanghai, China\\
\textsuperscript{\rm 3}The University of Hong Kong, Hong Kong, China, 
\textsuperscript{\rm 4}The Skolkovo  Institute  of Science and Technology, Moscow, Russia\\ 
\textsuperscript{\rm 5}Nagoya Institute of Technology, Nagoya, Japan; RIKEN Center for Advanced Intelligence Project, Japan\\      
\{shiqiquan, lc.leichen, yuan.mingxuan, zeng.jia\}@huawei.com,
14jiamingyin@tongji.edu.cn,  
jjcai@connect.hku.hk,\\
a.cichocki@skoltech.ru, 
t.yokota@nitech.ac.jp
}

\begin{document}

\maketitle

\begin{abstract}
This work proposes a novel approach for multiple time series forecasting. At first, multi-way delay embedding transform (MDT)  is employed to represent time series as low-rank block Hankel tensors (BHT). Then,  the higher-order tensors are projected to compressed core tensors by applying  Tucker decomposition. At the same time, the  generalized tensor Autoregressive Integrated Moving Average (ARIMA) is explicitly used on consecutive core tensors to predict future samples.  In this manner, the proposed approach tactically incorporates the unique advantages of MDT tensorization (to exploit mutual correlations) and tensor ARIMA coupled with low-rank Tucker decomposition into a unified framework.  This framework exploits the  low-rank structure of block Hankel tensors in the embedded space and captures the intrinsic correlations among multiple TS, which thus can improve the forecasting results, especially for multiple short time series.  Experiments conducted on three public  datasets and two industrial datasets verify that the proposed BHT-ARIMA effectively improves forecasting accuracy and reduces computational cost compared with the state-of-the-art methods.

\end{abstract}

\subsection{Introduction}
Time series forecasting (TSF)  is one of the most sought-after and yet arguably the most challenging tasks. 	It  has played an important  role in a wide range of areas including statistics, machine learning, data mining,  econometrics, operations research for several decades. 
	For example, forecasting the supply and demand of products can be used for  optimizing inventory management, vehicle scheduling and topology planning, which are    crucial for most aspects of supply chain optimization \cite{faloutsos2019forecasting}.
	
	Among existing TSF approaches, autoregressive integrated moving average (ARIMA) \cite{box1968some} is one of the most popular and widely used  linear  models due to its  statistical properties and great flexibility \cite{liu2016online}.  ARIMA  merges       the autoregressive model (AR) and the moving average model (MA)   with  differencing techniques for non-stationary TSF.   However, most existing ARIMA models need to predict multiples TS one by one and thus suffer from high computational cost, especially for a large number of TS.  In addition,   ARIMA models  do not consider  the intrinsic relationships among   correlated TS, which may limit their performance.  Recently, researchers from
	Facebook developed   Prophet \cite{taylor2018forecasting}.  It  can   estimate  each  TS  well based on an additive model where non-linear trends are fit with seasonality and holidays. 
	However, its computational expense grows steeply with increasing number of TS due to it estimates  single TS separately. 
	
	Multiple TS arising from real applications  can be reformulated as a matrix or even a high-order  tensor (multi-way data) naturally. For example,  the spatio-temporal grid of ocean data in meteorology can be shaped as  a fourth-order tensor TS, wherein four factors are jointly represented as latitude, longitude, grid points and time \cite{jing2018high}. When
	dealing with tensors, traditional linear TSF models require reshaping  TS into vectors. This vectorization not only  causes a loss of
	intrinsic structure information but also  results in  high computational and memory demands.
	
	Tensor decomposition is a powerful computational technique for extracting valuable information from tensorial data \cite{cichocki2016tensor,shi2018feature,zhou2019probabilistic,zhou2019bayesian}. Taking this advantage, tensor decomposition-based   approaches can handle  multiple   TS simultaneously and achieve good  forecasting performance \cite{dunlavy2011temporal,li2015tensor,tan2016short,bhanu2018forecasting,faloutsos2018forecasting}. 
	For example,   Tucker decomposition  integrated with   AR model was proposed   \cite{jing2018high}     to  obtain    the multilinear orthogonality  AR (MOAR) model and the multilinear constrained AR model for high-order TSF. Moreover, some works incorporate  decomposition with neural networks for more complex  tensorial TS \cite{yu2017long,ma2019large}. 
	
	However, instead of being high-dimensional tensors in nature, many real-world TS are relatively short and small \cite{smyl2016data}.
	For example,  since the entire   life  cycle of    electronic products like smartphones or laptops   is usually  quite short (around one year for each generation),  their  historical demand data of product materials and sales records are   very  small. Due to limited information,    the future demands or sales of these products cannot be effectively predicted  by \textit{either} existing linear models  \textit{or}   tensor methods  and  deep learning approaches.

	Multi-way delay embedding transform (MDT)  \cite{yokota2018missing}  is an emerging technique to Hankelize  available data to a high-order block Hankel tensor. In this paper, we conduct MDT on multiple  TS along the temporal direction, and thus get a  high-order block Hankel tensor   (\textit{i.e. a block tensor whose entries are Hankel  tensors \cite{ding2015fast}}),  which represents all the  TS at each time point    as a tensor  in a high-dimensional embedded space. The transformed  high-order data are assumed to have   a low-rank  or  smooth manifold  in the embedded space \cite{yokota2018missing,yokota2018tensor}.
	With the block Hankel tensor, we then employ low-rank Tucker decomposition  to  learn  compressed core tensors by    orthogonal factor (projection) matrices.  These  projection matrices  are jointly used  to maximally preserve the temporal continuity between core tensors which can better capture the intrinsic  temporal correlations than the original TS data. 
	At the same time,  we generalize classical ARIMA to tensor form and directly apply it on the core tensors \textit{explicitly}.  Finally, we predict a new core tensor at the next time point, and then obtain forecasting results for all the TS data simultaneously via inverse Tucker decomposition and  inverse MDT.  In short, we  incorporate block Hankel tensor  with ARIMA (\textbf{BHT-ARIMA}) via low-rank Tucker decomposition into  a unified framework.    This framework  exploits  low-rank data structure in the embedded space and captures the intrinsic correlations among multiple TS, leading to  good forecasting results. 
	We evaluate  BHT-ARIMA on  three public  datasets and two industrial datasets with  various settings. 
	In a nutshell,  \textbf{the main  contributions of this paper are threefold:}

	\begin{enumerate}
		\item [1)]  We  employ  MDT along the temporal direction to transform  multiple TS to a high-order block Hankel tensor.  To the best of our knowledge, we are the first  to introduce MDT together with tensor decomposition into  the field of TSF. In addition, we empirically demonstrate this strategy is also  effective for existing  tensor  methods. 
		\item  [2)]   We propose to apply the  orthogonal  Tucker decomposition   to explore  intrinsic   correlations among  multiple TS  represented as the compressed core tensors     in the low-rank embedded  space.  Furthermore, we empirically study  BHT-ARIMA with relaxed-orthogonality (\textit{impose the   orthogonality  constraint  on all the modes  \textbf{except}    the  temporal mode}) which can obtain even slightly better  results with reduced sensitivity to  parameters.  
		\item [3)]  We generalize the classical  ARIMA to tensor form and  incorporate it into   Tucker decomposition in a unified framework. We explicitly use the learned informative core tensors  to train the ARIMA model. Extensive experiments  on both public and industrial datasets validate and illustrate    that BHT-ARIMA significantly outperforms nine state-of-the-art (SOTA) methods in both efficiency and effectiveness, especially for multiple short TS.
	\end{enumerate}

\section {Preliminaries and Related Work}

\subsubsection{Notations}
The number of dimensions of a tensor is the \emph{order} and each dimension is a \emph{mode} of it.  A vector  is denoted by a bold lower-case letter $\mathbf{x} \in \mathbb{R}^I$.
A matrix is denoted by a bold capital letter $\mathbf X \in \mathbb{R}^{I_1 \hspace{-0.05cm}\times\hspace{-0.05cm} I_2 }$.
A higher-order ($N \geq 3$) tensor is denoted by a bold calligraphic letter $\ten{X}\in \mathbb{R}^{I_1\hspace{-0.05cm} \times I_2\hspace{-0.01cm} \cdots \hspace{-0.05cm}\times I_N}$.
The $i_{th}$ entry of  $\mathbf x$ is denoted by $\mathbf x_{i}$, and the $(i,j)$th entry of $\mathbf X$ is denoted by $ \mathbf X_{i_1,i_2}$.
The $(i_1, \ldots, i_N)$th entry of $\ten{X} $ is denoted by $\ten{X}_{i_1, \ldots, i_N}$.
The Frobenius norm of a tensor  $\ten X$ is defined by $\|\hspace{-0.05cm}\ten X\hspace{-0.05cm}\|_F \hspace{-0.05cm}= \hspace{-0.05cm} \sqrt{<\hspace{-0.05cm}\ten X, \ten X\hspace{-0.05cm} >}$, where  $< \hspace{-0.05cm}\ten{X}, \ten{X} \hspace{-0.05cm}> =\hspace{-0.1cm}\sum_{i_1}\hspace{-0.1cm}\sum_{i_2} \hspace{-0.05cm}\cdots\sum_{i_N}\hspace{-0.05cm} \ten{X}_{i_1, \ldots, i_N}^2$ denotes inner product.

A mode-$n$ product of  $\ten{X}$ and   $  \mathbf U\hspace{-0.1cm} \in\hspace{-0.1cm} \mathbb{R}^{I_n\hspace{-0.05cm} \times \hspace{-0.05cm}R_n}$   is denoted by  $ \ten{Y} \hspace{-0.05cm} = \hspace{-0.05cm} \ten{X} \hspace{-0.05cm} \times_n \hspace{-0.05cm} \mathbf U^\top \hspace{-0.09cm}\in \hspace{-0.09cm}\mathbb{R}^{I_1 \hspace{-0.05cm} \times \hspace{-0.025cm} \cdots \hspace{-0.025cm} \times \hspace{-0.05cm} I_{n-1} \hspace{-0.05cm} \times R_n \hspace{-0.03cm}  \times\hspace{-0.05cm}  I_{n+1} \hspace{-0.025cm}  \times \hspace{-0.025cm} \cdots \hspace{-0.025cm} \times I_N}$.
Mode-$n$ unfolding is the process of reordering the elements of a tensor	into matrices along each mode. A mode-$n$ unfolding matrix of  $\ten{X}$ is denoted as $\mathbf X^{(n)}  = \text{Unfold} (\ten{X})  \in \mathbb{R}^{I_n \times \Pi_{i \neq n} I_{i}  }$. Its inverse operation is fold: $\ten{X}  = \text{Fold} (\mathbf X^{(n)})$.

\subsubsection{Tucker Decomposition}
It represents 
a  tensor  $\ten{X}_t \in \mathbb{R}^{I_1 \times I_2  \times \cdots  \times I_N}$   as a core tensor with factor (projection) matrices \cite{kolda2009tensor}: 
\begin{equation}
\begin{aligned}
\small
\label{TuckerModel0}
\ten{X}_t &   =  &
\ten{G}_t {\times_1} \mathbf{U}^{(1)}  {\times_2}  \mathbf{U}^{(2)}  \cdots {\times_N} \mathbf{U}^{(N)},
\end{aligned}
\end{equation}
where $ \{\mathbf{U}^{(n)}  \in \mathbb{R}^{I_n \times  R_n}, n= 1,  \ldots, N, \ \text{and} \ R_n < I_n \}$
are projection  matrices which usually has orthonormal columns and $\ten{G}_t \in \mathbb{R}^{R_1 \times R_2 \times  \cdots \times R_N}$ is the core tensor with lower dimensions. 
The {Tucker-rank} of the  $\ten{X}_t$ is an $N$-dimensional vector: $[R_1, \hspace{-0.05cm} \ldots \hspace{-0.05cm}, \hspace{-0.05cm} R_n,  \hspace{-0.05cm}\ldots \hspace{-0.05cm}, R_N]$, where $n$-th entry $R_n$ is the rank of the mode-$n$ unfolded matrix $\mathbf{X}{_t^{(n)}}$.

\subsection{ARIMA}

Let  $\mathbf x_t$ as the actual data value at any time point  $t$.   $\mathbf x_t$ can be considered as a linear function of the past $p$  values and  past  $q$ observation of random errors, i.e., a   ARMA $(p,q)$ model:  
\begin{equation}\label{eq:arma} 
\small
\begin{aligned}
\mathbf x_t = \sum_{i=1}^{p}\alpha_i \mathbf x_{t-i} -  \sum_{i=1}^{q}  \beta_i \boldsymbol  \epsilon_{t-i}   +  \boldsymbol \epsilon_t,  
\end{aligned}  
\end{equation}
where  the  random errors $\{\boldsymbol \epsilon_t\}$ are identically distributed with a mean of zero and a constant variance. 
$\{\alpha_i\}_{i=1}^{p}$ and $ \{\beta_i \}_{i=1}^{q}$ are the coefficients  of AR and MA, respectively.  
In practice, TS data are usually not  stationary. The ARIMA model integrates a differencing method to deal with non-stationary TS data. 
Let  $\Delta^d \mathbf x_t $ denote as the  order-$d$  differencing  of $ \mathbf x_t$ and  an ARIMA$(p,d,q)$ model is given by:
\begin{equation}\label{eq:arima} 
\small 
\begin{aligned}
\Delta^d  \mathbf x_t = \sum_{i=1}^{p}\alpha_i \Delta^d  \mathbf x_{t-i} -  \sum_{i=1}^{q}  \beta_i \boldsymbol  \epsilon_{t-i}   +  \boldsymbol \epsilon_t.  
\end{aligned} 
\end{equation}
It  has been one of the most popular TSF models and has  many variants  \cite{zhang2003time,khashei2011novel,liu2016online}.  ARIMA models are usually used  for   single  TSF.

\begin{figure}[ttt!]
	\centering
	\subfloat[BHT-ARIMA for second-order tensor TS]{\includegraphics[ width=0.88\columnwidth]{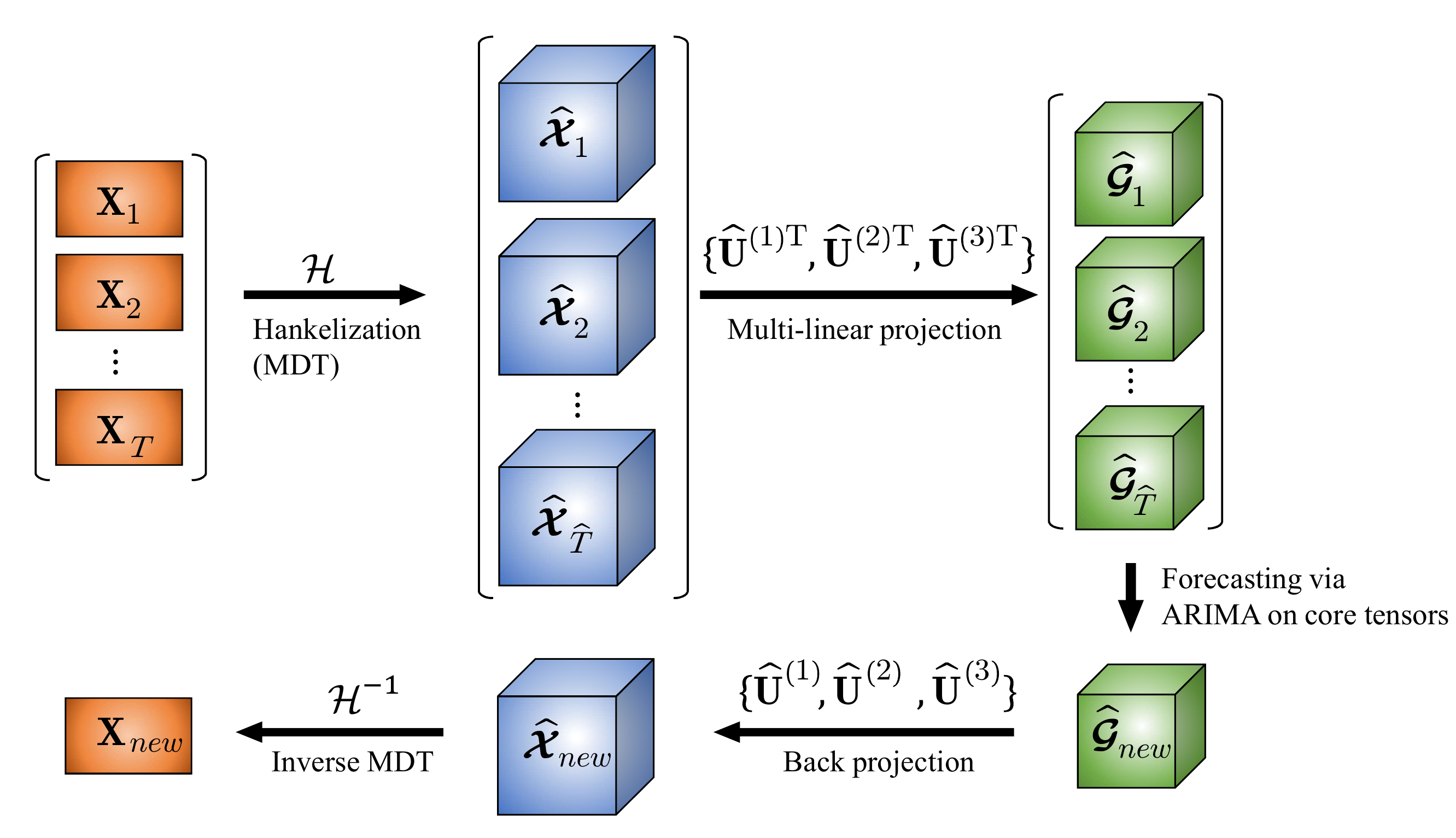}}  \label{BHTARIMA_2Dcase}
	\subfloat[BHT-ARIMA for third-order tensor TS ]{\includegraphics[ width=0.98\columnwidth]{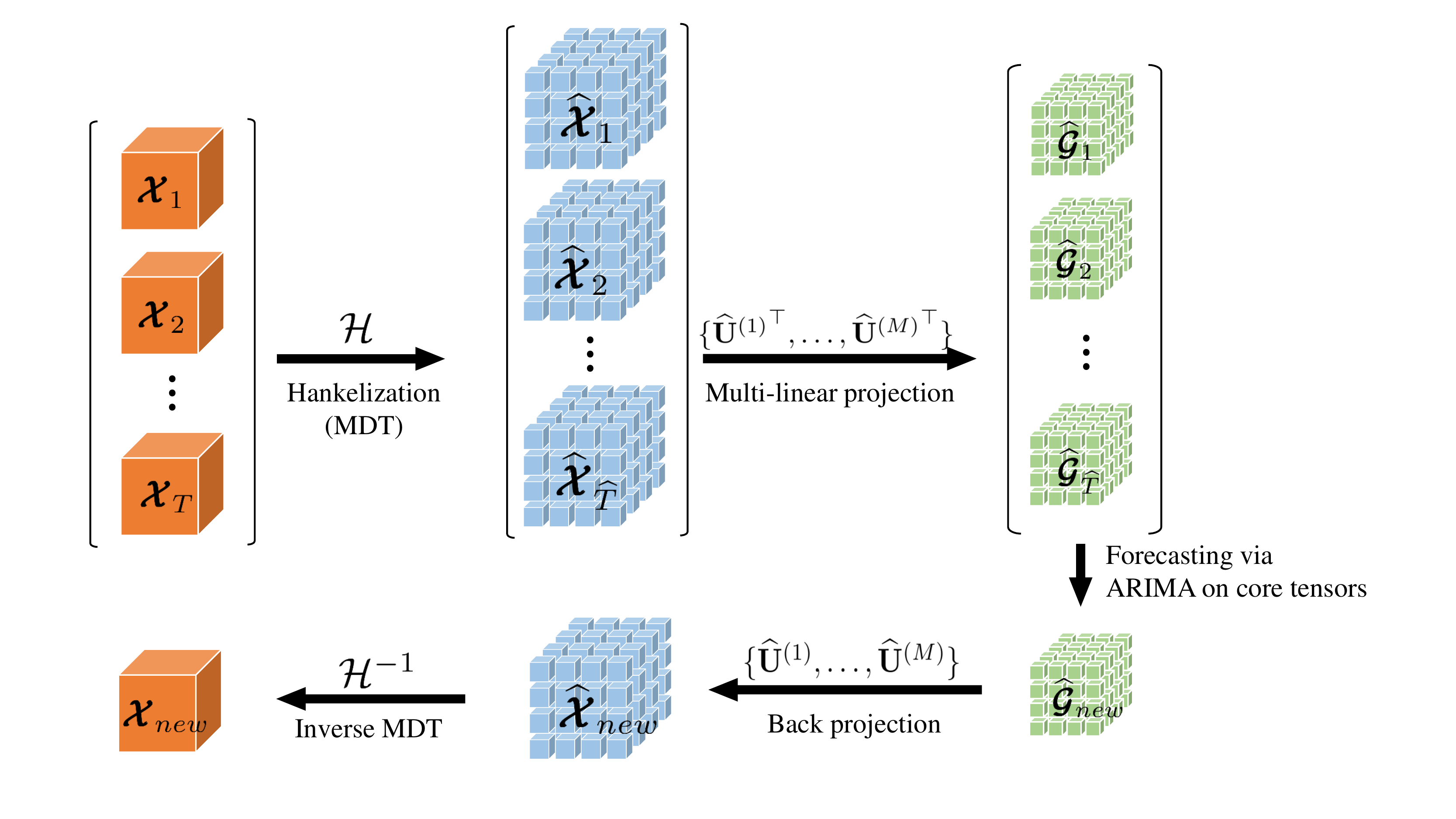}} \label{BHTARIMA_3Dcase}
	\caption{\label{BHTARIMAfigure}Illustration of the proposed method for prediction of multiple TS represented by a set of second/third-order tensors. Extension to higher-order tensors is straightforward.}
\end{figure}

\subsection{Tensor decomposition-based methods}

These tensor  approaches preserve the structure of high-order data and can handle multiple TS simultaneously \cite{rogers2013multilinear,fanaee2016tensor,de2017tensorcast,bhanu2018forecasting,agarwal2018model}. The closest  related work is the  MOAR which collectively integrates    Tucker decomposition  with AR   for high-order TSF \cite{jing2018high}:
\begin{equation}\label{MOAR}
\small
\begin{aligned}
\hspace{-0.1cm} \mathop{\min}_{ \{{\mathbf U}^{(i)}\}, \{\alpha_i \}  } &  \sum_{t=2}^{{T}}\ \Big \| \Big(  {{\ten{X}}_t}  -  \sum_{i=1}^{p} \alpha_i  {{\ten{X}}_{t-i}}  \Big) \prod_{n=1}^N \times_n { \mathbf U}{^{(n)}}^\top   \Big {\|} _F^2   \\&  \text{s.t.}  \ \ { \mathbf U}{^{(n)}}^\top { \mathbf U}{^{(n)}} = \mathbf I, n=1,\ldots, N, 
\end{aligned}
\end{equation}
where   $\mathbf I$ is an identity matrix and  the core tensors  are formulated in Tucker model and   \textit{implicitly}  used  to train the  AR model.  Our BHT-ARIMA  differs  from this closest related MOAR in three key aspects:1) BHT-ARIMA utilizes the BHT as input which can be better estimated than original data, especially for shorter univariate TS.   This will be also  verified by using BHT to improve MOAR;
2) BHT-ARIMA explicitly  uses low-rank core tensors to train its model while  MOAR does not directly use them. Since the core tensors are  smaller and can  better capture the intrinsic temporal correlations than  original TS data. BHT-ARIMA can improve computational speed  and   forecasting accuracy than MOAR;    
3)  BHT-ARIMA removes non-stationarity  of TS data by using  differencing technique  which is not considered in MOAR.

Besides, some works  integrate  decomposition with neural networks \cite{chen2018neucast,ma2019large,sun2019bayesian}, like tensor-train decomposition  with recurrent neural network (TTRNN) \cite{yu2017long}   can  achieve promising results for higher-order non-linear TS.   However, these tensor methods are not applicable for  short  univariate TS which cannot be presented as high-order tensors naturally.

\subsection{MDT for multiple  TS}
Hankelization is an effective way to transform  lower-order data to higher-order tensors.  MDT is a multi-way extension of Hankelization  and show good results in tensor completion  \cite{yokota2018missing,yokota2019manifold}. It combines  multi-linear duplication and multi-way folding operations.  By denoting $\widehat{\ten{X}}\hspace{-0.02cm}  \in \hspace{-0.02cm} \mathbb{R}^{J_1 \times\hspace{-0.02cm}  J_2 \hspace{-0.02cm} \cdots\hspace{-0.02cm}  \times\hspace{-0.02cm}  J_M}$ as the block Hankel tensor of   $\ten{X}\hspace{-0.02cm}   \in\hspace{-0.02cm}   \bbR{I_1 \hspace{-0.02cm}  \times \hspace{-0.02cm}  I_2 \hspace{-0.02cm}  \times\hspace{-0.02cm}   \cdots\hspace{-0.02cm}   \times \hspace{-0.02cm}  I_N}, N < M$, the  MDT  for  $\ten{X}$  is defined by
\begin{equation}\label{MDT}
\small 
\begin{aligned}
\widehat{\ten{X}} =\ten{H}_{\bm \tau}(\ten{X}) = \text{Fold}_{(\mathbf I, \bm \tau)} ( \ten{X} \times_1 \mathbf S_1 \cdots \times_N \mathbf S_N),
\end{aligned}
\end{equation}
where $\mathbf S_n\hspace{-0.02cm}  \in \hspace{-0.02cm}  \mathbb{R}^{\tau_n(I_n-\tau_n+1) \hspace{-0.02cm}  \times\hspace{-0.02cm}  I_n}$ is a duplication matrix  and  $\text{fold}_{(\mathbf I, \bm \tau)}\hspace{-0.02cm} :\hspace{-0.02cm}  \bbR{\tau_1(I_1-\tau_1+1) \times \cdots \times \tau_N(I_N-\tau_N+1)}$ $\rightarrow$ $\bbR{\tau_1 \times (I_1-\tau_1+1) \times \cdots \times \tau_N \times (I_N-\tau_N+1)}$ constructs a higher order block Hankel tensor $\widehat{ \ten{X}}$ from the input  tensor ${ \ten{X}}$.   The inverse MDT for  $\widehat{ \ten{X}}$ is given by 
\begin{equation} 
\begin{aligned}
{\ten{X}} = \mathcal{H}_{\bm \tau}^{-1}(\widehat{ \ten{X}}) = \text{Unfold}_{(\mathbf I, \bm \tau)}(\widehat{ \ten{X}}) \times_1 \mathbf S_1^\dagger \cdots \times_N \mathbf S_N^\dagger,
\end{aligned}
\end{equation}
where $\dagger$ is the  Moore-Penrose pseudo-inverse.

\section{Proposed Block Hankel Tensor ARIMA}\label{sec:proposed}
To  effectively and efficiently address the multiple (short) TSF problem,  
we  propose to  incorporate block Hankel tensor  with ARIMA (\textbf{BHT-ARIMA}) via low-rank Tucker decomposition. The main idea of the proposed method is illustrated in Fig. \ref{BHTARIMAfigure}. It consists of three major steps.

\begin{figure}[ttt!]
	\centering
	\includegraphics[width=0.9\columnwidth ]{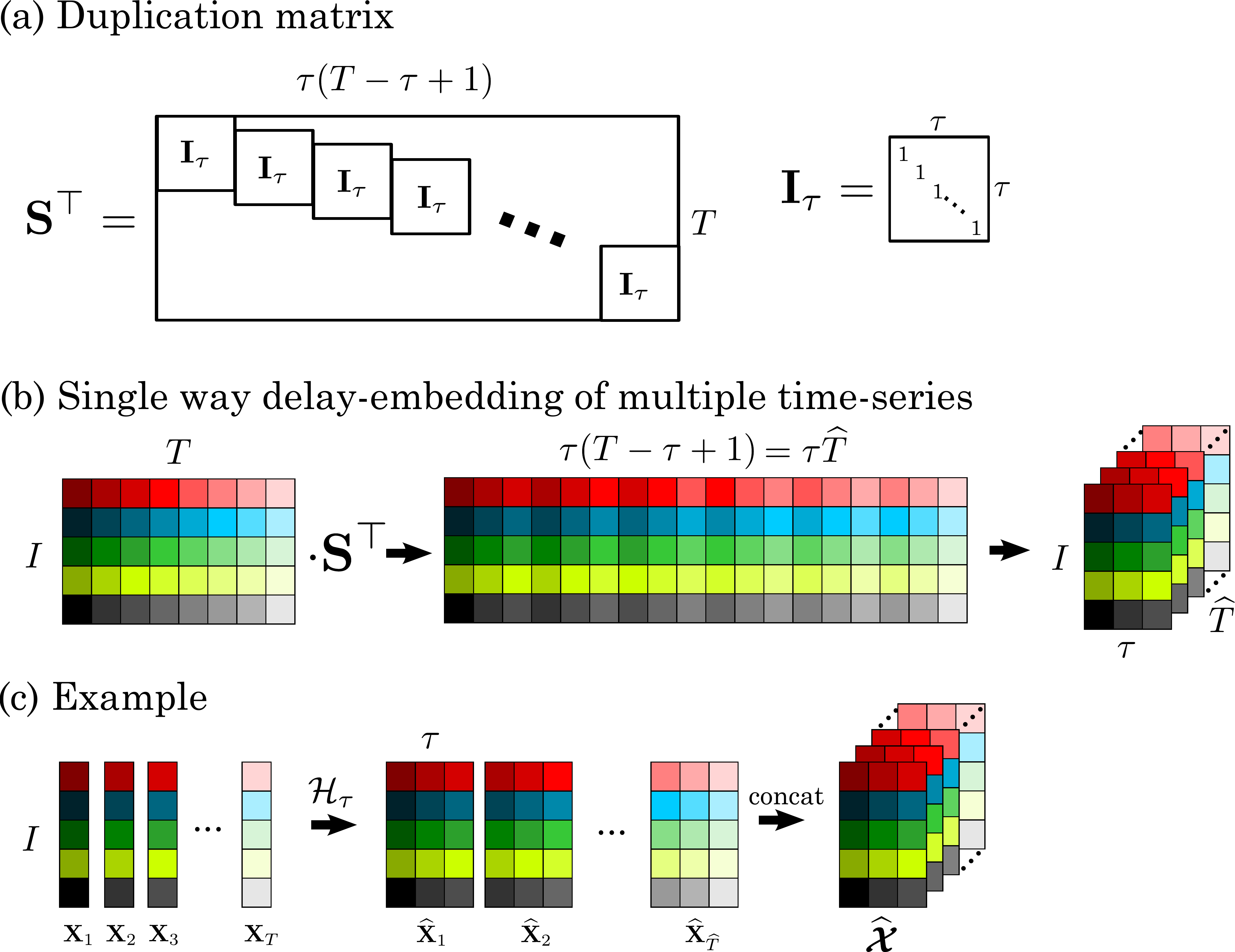}
	\caption{\label{MDTexample}Illustration of applying MDT along the temporal mode of multiple TS  represented  as a matrix $\mathbf{X} \in \bbR{I \times T}$. As we view the last mode $N$ ($N=2$ in this example) of multiple TS as the  temporal mode, i.e., $I_N =T$, the last mode  duplication matrix   $\mathbf S_N\hspace{-0.02cm}  \in \hspace{-0.02cm}  \mathbb{R}^{\tau_N (I_N-\tau_N+1) \hspace{-0.02cm}  \times\hspace{-0.02cm}  I_N} =  \mathbf S\hspace{-0.02cm}  \in \hspace{-0.02cm}  \mathbb{R}^{\tau(T-\tau+1) \hspace{-0.02cm}  \times\hspace{-0.02cm}  T}$. Note that  $\mathbf X \times_N \mathbf S$ =   $\mathbf X \mathbf S^\top$ for  matrix case.  Extension to multi-modes  is straightforward for tensors.}
\end{figure}

\subsection{Step 1: Block Hankel Tensor via  MDT} 

In this step, we  aim to employ MDT to transform multiple  TS  to  a   high-order block Hankel tensor.     Let $\ten{X} \in \bbR{I_1 \times \cdots \times I_N\times T}$ be the input  data where  each fiber  is one TS and  the last mode $T$ is the temporal mode.  We conduct the MDT only along the temporal direction, i.e.,  
\begin{equation}\label{MDTforX}
\begin{aligned}
&{\widehat{\ten{X}}} = \mathcal{H}_{\tau} (\ten{X}) \in \bbR{J_1 \times \cdots \times J_M \times \widehat{T}}, 
\end{aligned} 
\end{equation}
where $\hspace{-0.05cm} J_n\hspace{-0.05cm} =\hspace{-0.05cm}I_n\hspace{-0.05cm}$  for  $n\hspace{-0.05cm}=\hspace{-0.05cm}1, ...,N$, $J_M=\tau$ and  $\widehat{T} \hspace{-0.05cm} = \hspace{-0.05cm} T \hspace{-0.05cm} -\hspace{-0.05cm} \tau\hspace{-0.05cm}  + \hspace{-0.05cm}1$. In this way, we get a block Hankel tensor in   high-dimensional embedded space, where each frontal slice  ${\widehat{\ten{X}}_t} \in \bbR{J_1 \times \cdots \times J_M}$ contains  the data points of all the TS  at $t$-th time point. 

Fig. \ref{MDTexample} illustrates   conducting  MDT only along the temporal mode of a second-order tensor, i.e.,  using the duplication matrix  $\mathbf S$ with $\tau$ to transform   $\mathbf{X} \in \bbR{I \times T}$ to  a third-order block Hankel tensor $\widehat{\ten{X}  } \in \bbR{I \times  \tau \times {(T-\tau+1)} }$.    The  block Hankel tensor is assumed to be  low-rank  or smooth in the embedded space \cite{yokota2018missing}.

\paragraph{\textit{Remark 1:}}   We   only apply MDT along the temporal mode because   the   relationship  between neighbor  items  of multiple TS    is usually not  stronger than their temporal correlation.  Thus, it is unnecessary to conduct MDT on all the modes   which probably not be meaningful while costing  more time (we  empirically study it, see Fig. 6 in the Supplementary). Nevertheless, our proposed method is capable to apply MDT on more modes to get a higher-order block Hankel tensor.

In this paper, we mainly handle  second-order and third-order original tensors, see more details of applying MDT on them in Appendix A.1 of Supplementary Material (\textbf{Supp.})\footnote{Available at \url{https://github.com/yokotatsuya/BHT-ARIMA}.}.

\subsection{Step 2: Tensor ARIMA with Tucker Decomposition}
In this step,  we generalize the classical  ARIMA to tensor form and  incorporate it into  Tucker decomposition.  Specifically, with the block Hankel tensor,  we  compute its order-$d$ differencing   to  get $\{ \Delta^d  \widehat{\ten{X}}_t\}_{t=d}^{\widehat{T}}$.
We then  employ  Tucker decomposition for    $\{ \Delta^d  \widehat{\ten{X}}_t\}$ by  projecting  it  to  core tensors  $ \{ \Delta^d  \widehat{\ten{G}}_t\}$ using  joint orthogonal factor matrices $ \{\widehat{\mathbf U}{^{(m)}}\}$, i.e.,  
\begin{equation}
\begin{aligned}\label{eq:Tucker_com0} 
\small
\Delta^d & {\widehat{\ten{G}}_t}      =  \Delta^d  {\widehat{\ten{X}}_t} \hspace{-0.1cm} \times_1   \widehat{\mathbf U}{^{(1)}}^\top \cdots \hspace{-0.05cm}  \times_M \widehat{\mathbf U}{^{(M)}}^\top   \\& \text{s.t.}  \ \ \widehat{ \mathbf U}{^{(m)}}^\top \widehat{ \mathbf U}{^{(m)}} = \mathbf I, m=1,\ldots, M, 
\end{aligned}
\end{equation}
where the projection matrices $ \{\widehat{\mathbf U}{^{(m)}} \in \bbR{J_m \times R_m} \}$  maximally preserve the temporal continuity between core tensors and the low-rank core tensors $ \{{ \Delta^d  \widehat{\ten{G}}_t}\in \bbR{R_1 \times \cdots \times R_M}\}$  represent the most important information of original Hankel tensors and reflect the intrinsic interactions between  TS.

Then, it would be more promising to train a good forecasting model  directly using   core tensors  \textit{explicitly} instead of whole  tensors. To   retain the temporal correlations  among core tensors, we generalize  a $(p,d,q)$-order  ARIMA model to tensor form and use it to   connect  the current core tensor ${ \Delta^d  \widehat{\ten{G}}_t} $ and the previous core tensors
${\Delta^d  \widehat{\ten{G}}_{t-1}},\Delta^d  {\widehat{\ten{G}}_{t-2}}, \ldots, \Delta^d  {\widehat{\ten{G}}_{t-p}}$ as follows:
\begin{equation} \label{TARIMA}
\begin{aligned}
\Delta^d  {\widehat{\ten{G}}_{t}}    & =  \sum_{i=1}^{p} \alpha_i  {\Delta^d   \widehat{\ten{G}}_{t-i}}  - \sum_{i=1}^{q} \beta_i {\widehat{\ten{E}}_{t-i}} + \widehat{\ten{E}}_t, 
\end{aligned}
\end{equation}
where the  $\{\alpha_i \}$ and  $\{\beta_i\}$  are the coefficients of AR and MA, respectively, and $\{\widehat{\ten{E}}_{t-i}\}$ are the random errors of past $q$ observations.  In model \eqref{TARIMA}, $\widehat{\ten{E}}_t$ is the forecast error at the current time point, which should be minimized  to optimal  zero. We  thus can derive the following  objective function:
\begin{equation}\label{HTARIMA_XG1}
\small 
\begin{aligned}
&\hspace{-0.4cm} \mathop{\min}_{ \{\widehat{\ten{G}}_t,\hspace{-0.03cm} \widehat{\mathbf U}{^{(m)}}, \hspace{-0.01cm}  \widehat{\ten{E}}_{t-i}, \hspace{-0.03cm} \alpha_i, \hspace{-0.03cm} \beta_i  \}}   \hspace{-0.05cm}\sum_{t=s+1}^{\widehat{T}}\hspace{-0.05cm}  \Bigg (\hspace{-0.05cm} \frac{1}{2} \Big \|  \hspace{-0.05cm}  {\Delta^d   \widehat{\ten{G}}_t} \hspace{-0.07cm}  - \hspace{-0.07cm} \sum_{i=1}^{p} \hspace{-0.03cm} \alpha_i \hspace{-0.03cm} {\Delta^d  \hspace{-0.01cm}  \widehat{\ten{G}}_{t-i}}   \hspace{-0.06cm}  + \hspace{-0.05cm}  \sum_{i=1}^{q}  \hspace{-0.05cm}  \beta_i  \hspace{-0.03cm}  {\widehat{\ten{E}}_{t-i}} \hspace{-0.03cm}  \Big {\|} _F^2  \\ & \ \ \ \ \ \ \ \ \ \ \ \ \ \ \ \  + \frac{1}{2}  \Big \| {\Delta^d   \widehat{\ten{G}}_t}    - {\Delta^d  \widehat{\ten{X}}_t}  \times_1  \widehat{\mathbf U}{^{(1)}}^\top   \cdots \times_M  \widehat{\mathbf U}{^{(M)}}^\top \Big  \| _F^2 \hspace{-0.05cm}  \Bigg)  \ \ \\ &  \ \ \ \ \ \ \ \ \ \ \   \ \ \ \ \ \ \ \  \  \text{s.t.}  \ \ \widehat{ \mathbf U}{^{(m)}}^\top \widehat{ \mathbf U}{^{(m)}} = \mathbf I, m=1,\ldots, M, 
\end{aligned}
\end{equation}
where $s\hspace{-0.05cm} = \hspace{-0.05cm}p\hspace{-0.05cm}+\hspace{-0.05cm}d\hspace{-0.05cm}+\hspace{-0.05cm}q$ is the sum of  ARIMA orders, and  is also the minimum input length of each  TS.
Next, we  solve this problem using augmented Lagrangian  methods. To facilitate the derivation of \eqref{HTARIMA_XG1}, we reformulate the optimization problem by unfolding each tensor variable along mode-$m$:
\begin{equation}\label{HTARIMA_obj} 
\small 
\begin{aligned}&  
\hspace{-0.1cm}\mathop{\min}_{\{{\widehat{\mathbf{G}}^{(m)}_{t}},  {\widehat{\mathbf{E}}^{(m)}_{{t-i}}},     \widehat{\mathbf U}{^{(m)}},\alpha_i,\beta_i\}      }    \sum_{t=s+1}^{\widehat{T}} \sum_{m=1}^{M}    \Bigg (  \frac{1}{2}  \Big \| \Delta^d   {\widehat{\mathbf{G}}^{(m)}_{t}}  -   \sum_{i=1}^{p} \alpha_i  \Delta^d  {\widehat{\mathbf{G}}^{(m)}_{{t-i}}}  \\ &  \ \ \ \ \ \ \ \  +  \sum_{i=1}^{q} \beta_i {\widehat{\mathbf{E}}^{(m)}_{{t-i}}}   \Big {\|} _F^2  +  \frac{1}{2}   \Big  \|  \Delta^d  {\widehat{\mathbf{G}}^{(m)}_{t}}    -   \widehat{\mathbf U}{^{(m)}}^\top {\widehat{\mathbf{X}}^{(m)}_{t}} \widehat{\mathbf U}{^{(-m)}}^\top \Big  \| _F^2\Bigg )    \\ & \ \ \ \ \ \ \ \ \ \ \ \  \ \text{s.t.}  \ \ \widehat{ \mathbf U}{^{(m)}}^\top \widehat{ \mathbf U}{^{(m)}} = \mathbf I, m=1,\ldots, M, 
\end{aligned} 
\end{equation}
where  $\widehat{\mathbf U}{^{(-m)}} = \widehat{ \mathbf U}{^{(M)}}^\top  \otimes \cdots \widehat{ \mathbf U}{^{(m+1)}}^\top  \otimes  \widehat{ \mathbf U}{^{(m-1)}}^\top  \otimes \cdots \widehat{ \mathbf U}{^{(1)}}^\top   \in \mathbb{R}^{\prod_{j \neq m}R_j \times \prod_{j \neq m}I_j }  $. In the following, we can update each target variable using closed-form solutions \footnote{The detailed derivation is presented in  \textit{Appendix A of the Supp.}.}.

\subsubsection{Update  $ \widehat{\mathbf{G}}^{(m)}_{t}   $  }

Equation \eqref{HTARIMA_obj}   with respect to $ \Delta^d \widehat{\mathbf{G}}^{(m)}_{t}  $   is:
\begin{equation}\label{HTARIMA_G1}
\small 
\begin{aligned}  
\hspace{-0.1cm} \mathop{\min}_{\{ \widehat{\mathbf{G}}^{(m)}_{t}  \}}  
\sum_{t=s+1}^{\widehat{T}} \sum_{m=1}^{M}   & \Bigg ( \frac{1}{2}  \Big \| \Delta^d   {\widehat{\mathbf{G}}^{(m)}_{t}}  -   \sum_{i=1}^{p} \alpha_i  \Delta^d  {\widehat{\mathbf{G}}^{(m)}_{{t-i}}} +   \sum_{i=1}^{q} \beta_i {\widehat{\mathbf{E}}^{(m)}_{{t-i}}}   \Big {\|} _F^2    \\ &  +  \frac{1}{2}  \Big  \|  \Delta^d  {\widehat{\mathbf{G}}^{(m)}_{t}}    -   \widehat{\mathbf U}{^{(m)}}^\top {\widehat{\mathbf{X}}^{(m)}_{t}} \widehat{ \mathbf U}{^{(-m)}}^\top \Big  \| _F^2\Bigg ).  
\end{aligned}\
\end{equation}

Computing  the partial derivation of this cost function  with respect to ${\widehat{\mathbf{G}}^{(m)}_{t}} $ and equalize it  to zero, we  update $\widehat{\mathbf{G}}^{(m)}_{t}  $ by:
\begin{equation}\label{HTARIMA_UpdateG}
\small 
\begin{aligned}
\Delta^d  \widehat{\mathbf{G}}^{(m)}_{t}  \hspace{-0.1cm} & = \hspace{-0.1cm}  \frac{1}{2} \Big ( \widehat{\mathbf U}{^{(m)}}^\top {\widehat{\mathbf{X}}^{(m)}_{t}} \widehat{ \mathbf U}{^{(-m)}}^\top   \hspace{-0.05cm} \\ &  + \hspace{-0.05cm} \sum_{i=1}^{p} \alpha_i  \Delta^d  {\widehat{\mathbf{G}}^{(m)}_{{t-i}}}   \hspace{-0.05cm}- \hspace{-0.05cm}  \sum_{i=1}^{q} \beta_i {\widehat{\mathbf{E}}^{(m)}_{{t-i}}}            \Big).
\end{aligned}
\end{equation}

\subsubsection{Update $ \widehat{ \mathbf U}{^{(m)}} $  }
Equation \eqref{HTARIMA_obj} with respect to $ \widehat{ \mathbf U}{^{(m)}} $   is: 
\begin{equation}\label{HTARIMA_updateU0}
\small 
\begin{aligned}
\hspace{-0.1cm} \mathop{\min}_{ \{\widehat{ \mathbf U}{^{(m)}} \} }  \hspace{-0.05cm}  \sum_{t=s+1}^{\widehat{T}}  \hspace{-0.01cm} \sum_{m=1}^{M}  &    \frac{1}{2}  \Big \|   \Delta^d  {\widehat{\mathbf{G}}^{(m)}_{t}}  \hspace{-0.05cm}   -   \hspace{-0.05cm} \widehat{\mathbf U}{^{(m)}}^\top  \hspace{-0.05cm}  {\widehat{\mathbf{X}}^{(m)}_{t}}  \widehat{ \mathbf U}{^{(-m)}}^\top  \hspace{-0.05cm}  \Big  \| _F^2  
\\ & \text{s.t.}  \ \ \widehat{ \mathbf U}{^{(m)}}^\top \widehat{ \mathbf U}{^{(m)}} = \mathbf I, m = 1,\ldots, M.
\end{aligned}
\end{equation}

The minimization of \eqref{HTARIMA_updateU0} over  $\widehat{ \mathbf U}{^{(m)}}$  with orthonormal columns is equivalent to the maximization of the well-known
orthogonality  Procrustes problem \cite{higham1995matrix}, whose global optimal solution is,
\begin{equation} \label{HTARIMA_UpdateU}
\small 
\begin{aligned}
\widehat{ \mathbf U}{^{(m)}}   = \widehat{ \mathbf U^*}{^{(m)}} (\widehat{ \mathbf V^*}{^{(m)}}  )^\top,
\end{aligned} 
\end{equation} 
where $\hspace{-0.05cm}\widehat{ \mathbf U^*}{^{(m)}}\hspace{-0.05cm}$ and $ \hspace{-0.05cm}\widehat{ \mathbf V^*}{^{(m)}}  \hspace{-0.05cm}$
are the left and right singular vectors
of SVD of $ \scriptsize \hspace{-0.01cm}\sum_{t=s+1}^{\widehat{T}}  \hspace{-0.05cm} {\widehat{\mathbf{X}}^{(m)}_{t}}  \hspace{-0.01cm}    \widehat{ \mathbf U}{^{(-m)}}^\top \hspace{-0.1cm} \Delta^d  \hspace{-0.01cm} {\widehat{\mathbf{G}}{^{(m)}_{t}}}^\top \hspace{-0.05cm}$, respectively.

\paragraph{\textbf{Discussion 1}: \textit{Relaxed-orthogonality}}
We empirically explored the effect of  relaxing the  full-orthogonality  in  \eqref{HTARIMA_updateU0} by  \textbf{removing} the orthogonal constraints along the last  mode (viewed as the temporal mode of each $\widehat{\ten{X}}_t$ in the embedded space). This  strategy   probably  relaxes  the heavy constraints on temporal smoothness and thus would make the proposed model more flexible and  robust  to  variability of  parameters, observed from  our experimental results.  We relax the last mode $\widehat{\mathbf U}^{(M)}$ without orthogonality constraint,  and then compute the partial derivation of  Eq. \eqref{HTARIMA_updateU0} with respect to $\widehat{\mathbf U}^{(M)}$  and  equalize it to zero. Thus, we can update it by 
\begin{equation}\label{HTARIMA_UpdateUM}
\small 
\begin{aligned}
\widehat{ \mathbf U}{^{(M)}}        = &  \Big(\hspace{-0.01cm}  \sum_{t=s+1}^{\widehat{T}} {\widehat{\mathbf{X}}^{(M)}_{t}}      \widehat{ \mathbf U}{^{(-M)}}^\dagger  ({\widehat{\mathbf{X}}^{(M)}_{t}}      \widehat{ \mathbf U}{^{(-M)}}^\dagger)^\top  \hspace{-0.01cm}  \Big)^{-1}  \\&   \Big(\hspace{-0.01cm}  \sum_{t=s+1}^{\widehat{T}} {\widehat{\mathbf{X}}^{(M)}_{t}}      \widehat{ \mathbf U}{^{(-M)}}^\dagger \Delta^d  {\widehat{\mathbf{G}}{^{(M)}_{t}}}^\top   \hspace{-0.01cm}   \Big).
\end{aligned}
\end{equation}

\subsubsection{Update  $\widehat{\mathbf{E}}^{(m)}_{{t-i}} $  }

Equation  \eqref{HTARIMA_obj}   with respect to $\widehat{\mathbf{E}}^{(m)}_{{t-i}} $   is: 
\begin{equation} \label{HTARIMA_UpdatedE0} 
\small 
\begin{aligned}   
\hspace{-0.2cm}\mathop{\min}_{\{\widehat{\mathbf{E}}^{(m)}_{{t-i}}\} } \hspace{-0.05cm}  \sum_{t=s+1}^{\widehat{T}} \hspace{-0.05cm}  \sum_{m=1}^{M}    \hspace{-0.05cm}  \frac{1}{2}  \Big \| \hspace{-0.05cm} \Delta^d   {\widehat{\mathbf{G}}^{(m)}_{t}}  \hspace{-0.1cm} -  \hspace{-0.1cm}  \sum_{i=1}^{p} \alpha_i  \hspace{-0.05cm}  \Delta^d  {\widehat{\mathbf{G}}^{(m)}_{{t-i}}} \hspace{-0.1cm} +\hspace{-0.1cm}   \sum_{i=1}^{q} \beta_i {\widehat{\mathbf{E}}^{(m)}_{{t-i}}}  \hspace{-0.05cm}  \Big {\|} _F^2.  
\end{aligned}  
\end{equation}

Computing  the partial derivation of  Eq. \eqref{HTARIMA_UpdatedE0}  with respect to $\widehat{\mathbf{E}}{^{(m)}_{{t-i}}} $ and equalize it  to zero, 
we can update  $\widehat{\mathbf{E}}{^{(m)}_{{t-i}}} $ by  
\begin{equation} \label{HTARIMA_UpdatedEF}
\small 
\begin{aligned}  \hspace{-0.2cm}
\widehat{\mathbf{E}}{^{(m)}_{{t-i}}} \hspace{-0.04cm}  = \hspace{-0.04cm} \frac{\sum_{t=s+1}^{\widehat{T}}   \hspace{-0.1cm} 
	\Big( \hspace{-0.05cm} \Delta^d \hspace{-0.03cm} {\widehat{\mathbf{G}}^{(m)}_{t}}  \hspace{-0.1cm}  - \hspace{-0.1cm}   \sum_{i=1}^{p}\hspace{-0.05cm}  \alpha_i \Delta^d \hspace{-0.03cm}  {\widehat{\mathbf{G}}^{(m)}_{{t-i}}}   \hspace{-0.05cm} + \hspace{-0.05cm}  \sum_{j
		\neq i}^{q}  \hspace{-0.05cm}  \beta_j \hspace{-0.05cm}  \widehat{\mathbf{E}}{^{(m)}_{{t-j}}}\hspace{-0.05cm}  \Big)} 
{(s+1-\widehat{T}) \beta_i   }.
\end{aligned} 
\end{equation}

\subsubsection{Update  $ \{\alpha_i \},  \{\beta_i \} $  }
Regarding the coefficient parameters $ \{\alpha_i \}_{i=1}^p,  \{\beta_i \}_{i=1}^q $ of AR and MA  respectively in the objective function \eqref{HTARIMA_obj}, we follow the classical ARIMA  based on  Yule-Walker method  to estimate them.  We generalize a least squares modified Yule-Walker technique to support tensorial data  and  then estimate  the  $ \{\alpha_i \}_{i=1}^p $  from the core tensors. Then, we estimate the $\{\beta_i \}_{i=1}^q $  from the residual time series.

\begin{algorithm}[ttt!]	
	\small
	\caption{TSF using \textbf{BHT-ARIMA}} \label{BHTARIMAalg}	
	\begin{algorithmic}[1]
		\STATE {\bf  Input:}   A   time series data  $\ten{X}\in \bbR{I_1 \times \cdots \times I_N \times T}$,  $(p,d,q)$,   $\tau$,  maximum iteration $K$,   and  stop criteria  $\textit{tol}$.
		\STATE \textbf{\underline{Step 1: Block Hankel Tensor via MDT}}
		\STATE Use MDT to transform the original tensor as {a  block Hankel tensor}: ${\widehat{\ten{X}}} = \mathcal{H}_\tau (\ten{X}) \in \bbR{J_1 \times \cdots \times J_M \times \widehat{T}}$, where each frontal slice  ${\widehat{\ten{X}}_t}  \in \bbR{J_1 \times \cdots \times J_M}$ refers to all the time series at $t$-th time point. 
		\STATE Set	values for  $ 	
		\{R_m\}_{m=1}^{M}$ and initialize $\{\widehat{ \mathbf U}{^{(m)}}\}_{m=1}^{M}$ randomly. 	
		\STATE   \textbf{\underline{Step 2:   Tensor ARIMA with Tucker decomposition}}. 
		
		\STATE Conduct order-$d$ differencing  for  $\hspace{-0.01cm} \{\hspace{-0.01cm}{\widehat{\ten{X}}_t}\hspace{-0.01cm}\}_{t=1}^{\widehat{T}}\hspace{-0.01cm}$ and get  $ \{\hspace{-0.01cm}\Delta^d\hspace{-0.01cm} {\widehat{\ten{X}}_t}\}_{t=d+1}^{\widehat{T}}\hspace{-0.01cm}$  
		\STATE Initialize  $\{\{\widehat{\ten{E}}_{{t-i}} \in \bbR{R_1 \times \cdots \times R_M} \}_{i=1}^q\}_{i=s+1}^{\widehat{T}}$ randomly.
		\STATE\textbf{for}  {$k = 1, ...,K$}
		\STATE \quad Compute all the latent low-dimensional core tensors  \quad  $ \{ \Delta^d {\widehat{\ten{G}}_t}    = \Delta^d {\widehat{\ten{X}}_t}  \times_1  \widehat{\mathbf U}{^{(1)}}^\top   \cdots \times_M  \widehat{\mathbf U}{^{(M)}}^\top\}_{t=1}^{\widehat{T}}$;
		\STATE  \quad  Estimate coefficients  $ \{\alpha_i \}_{i=1}^p,  \{\beta_i \}_{i=1}^q $  of  AR and MA  \hspace{0.5cm} based on Yule-Walker equations based on $ \{\Delta^d {\widehat{\ten{G}}_t}\}_{t=d+1}^{\widehat{T}}$
		\STATE \quad \textbf{for} $n = 1, ...,M$
		\STATE	\quad \quad   Update $ \hspace{-0.05cm} \Delta^d\hspace{-0.05cm}  {\widehat{\mathbf{G}}^{(m)}_{t}}\hspace{-0.05cm}$  by \eqref{HTARIMA_UpdateG} and get $\Delta^d {\widehat{\ten{G}}_{t}} \hspace{-0.05cm}=\hspace{-0.05cm}  \text{Fold} \Big
		(\Delta^d{\widehat{\mathbf{G}}^{(m)}_{t}}\Big)  $ 							
		\STATE	\quad \quad   Update $\widehat{ \mathbf U}{^{(m)}}$ by \eqref{HTARIMA_UpdateU}
		
		\STATE \quad 	\quad  {If applying relaxed-orthogonality, update $\widehat{ \mathbf U}{^{(M)}}$  by \eqref{HTARIMA_UpdateUM}}. 
		\STATE \quad \quad \textbf{for} \quad  {$i = 1, ...,q$}	
		\STATE \quad\quad \quad  Update $\widehat{\mathbf{E}}^{(n)}_{{t-i}}$ by \eqref{HTARIMA_UpdatedEF} and update $ \widehat{\ten{E}}_{{t-i}}$ by $ \text{Fold} \Big
		(\widehat{\mathbf{E}}^{(m)}_{{t-i}}\Big)  $
		\STATE \quad    Convergence checking:  if  $\frac{ \sum_{m=1}^{M} ||\widehat{ \mathbf U}{^{(m)}}^{k+1}  -\widehat{ \mathbf U}{^{(m)}}^{k} ||_F^2} {\sum_{m=1}^{M}||\widehat{ \mathbf U}{^{(m)}}^{k+1}||_F^2 } < tol$,  break; otherwise, continue.
		\STATE {\underline{\textbf{Step 3: Forecasting}}}   
		\STATE  Compute new observation by \eqref{Prediction0} to get $ \Delta^d  {\widehat{\ten{G}}_{t+1}}$
		and then compute $\Delta^d  {\widehat{\ten{X}}_{\widehat{T}+1}} =  \Delta^d  {\widehat{\ten{G}}_{{t}+1}}   \prod_{m=1}^M \times_m \widehat{ \mathbf U}{^{(m)}} $. 
		\STATE  Conduct inverse   differencing  for   $ \Delta^d  {\widehat{\ten{X}}_{\widehat{T}+1}}$ and   get  $ {\widehat{\ten{X}}_{\widehat{T}+1}}$.
		\STATE Conduct  inverse MDT:  ${{\ten{X}}_{T+1}} = \mathcal{H}_\tau^{-1}( {{\widehat{\ten{X}}}_{\widehat{T}+1}})$
		\STATE {\bf Output:} ${{\ten{X}}_{T+1}}$, $\{\widehat{ \mathbf U}{^{(1)}}, \ldots, \widehat{\mathbf U}{^{(m)}}, \ldots, \widehat{ \mathbf U}{^{(M)}}\}$.
	\end{algorithmic} 
\end{algorithm}

\subsection{Step 3: Forecasting   ${{\ten{X}}_{T+1}}  $  }

Finally, we apply the learned model to get a new  $\Delta^d  {\widehat{\ten{G}}_{\widehat{T}+1}}$:
\begin{equation}\label{Prediction0}
\small 
\begin{aligned}
\Delta^d  {\widehat{\ten{G}}_{\widehat{T}+1}} =    \sum_{i=1}^{p} \alpha_i   \Delta^d  {\widehat{\ten{G}}_{\widehat{T}-i}}     - \sum_{i=1}^{q} \beta_i {\widehat{\ten{E}}_{\widehat{T}-i}}. 
\end{aligned}
\end{equation}

After obtaining $ \Delta^d  {\widehat{\ten{G}}_{\widehat{T}+1}}$, we reconstruct a new tensor by Tucker model with optimized factor matrices:  
$\Delta^d  {\widehat{\ten{X}}_{\widehat{T}+1}} =  \Delta^d  {\widehat{\ten{G}}_{{t}+1}}   \prod_{m=1}^M \times_m  \widehat{ \mathbf U}{^{(m)}} $. We then  
conduct inverse order-$d$  differencing  for   $ \Delta^d  {\widehat{\ten{X}}_{\widehat{T}+1}}$ and get   $ {\widehat{\ten{X}}_{\widehat{T}+1}}\in \bbR{J_1 \times \cdots \times J_M \times {(\widehat{T}+1)}} $  in the embedded space. Finally,  we  apply  inverse MDT to get ${{\ten{X}}_{T+1}}  = \mathcal{H}_\tau^{-1}({\widehat{\ten{X}}_{\widehat{T}\hspace{-0.05cm}+\hspace{-0.05cm}1}})   \in \bbR{I_1 \times \cdots \times I_N \times {(T+1)}} $ to get the predicted values at the  $T+1$-th time point for all TS \textit{simultaneously}. Furthermore, we could do long-term forecasting by using prior foretasted values in last steps. Although this way may lead to error accumulation to some degree, it becomes more practical for real-world applications \cite{jing2018high}. 

\paragraph{\textit{Remark 2:}}   We  explicitly use the compressed core tensors to train the model including   parameters estimation. That is different from existing tensor methods  like MOAR which implicitly use the core tensors to train their models.  In this manner, we not only reduce the computational cost since the size of core tensors $\{\widehat{\ten{G}}_{t} \} $  are much smaller than whole tensors $\{\widehat{\ten{X}}_{t} \} $ based on the low-rank assumption in the embedded space, but also  improve the forecasting accuracy by  utilizing the mutual correlations  among multiple TS in the model building process.  

Finally, we summarize
the proposed \textbf{BHT-ARIMA} in \textbf{Algorithm} \textbf{\ref{BHTARIMAalg}} and further evaluate it in the following section.

\section{Experiments}
Due to limited space,  we  present the detailed experimental setup and  parameter analysis  with   figures  in the  Supp.. 
\subsection{Experimental Setup}

\subsubsection{Datasets}
We  evaluate the proposed BHT-ARIMA by conducting  experiments on five real-world datasets, including: i)\textit{Three publicly available datasets}:
\textbf{Traffic}  is originally collected from California department of transportation   and describes the road occupy rate of Los Angeles County highway network. We here use the same subset  in  \cite{yu2017spatio} which selects 228 sensors randomly  and we aggregate it to daily interval  for each TS with  80 days data points.  
\textbf{Electricity}  records 321 clients' hourly electricity consumption \cite{lai2018modeling}. We merge the  every 24 time points to obtain  a daily interval TS dataset with size $321  \hspace{-0.05cm}\times \hspace{-0.05cm} 1096$.
\textbf{Smoke Video} records the Smoke  from the chimney of a factory taken in Hokkaido in 2007. We sample  and resize the images to obtain a  third-order TS dataset of size 36$  \hspace{-0.05cm}\times  \hspace{-0.05cm}$64$  \hspace{-0.05cm}\times  \hspace{-0.05cm}$100;  \\
ii)\textit{Two industrial datasets} from the supply chain  of   \textbf{Huawei}:
\textbf{PC sales} has  105 weekly sales records of 9 personal computers from 2017 to  2019; \textbf{Raw materials} includes  2246  material  items of   making a product, each item has 24 monthly demand quantities. 
We illustrate these datasets in Fig. $1$  and  $2$ in the {Supp..}

\subsubsection{Compared methods}  
We  compared \textit{nine} competing methods:  i) the   classical 
\textbf{ARIMA}, Vector AR (\textbf{VAR})  and \textbf{XGBoost}  \cite{chen2016xgboost}; ii) the two popular industrial forecasting methods: Facebook-\textbf{Prophet},  and Amazon-\textbf{DeepAR} \cite{salinas2017deepar}; iii) Neural network based methods\footnote{We didn't show the  comparison to Long Short-Term Memory (LSTM) as it yields similar results with GRU.}:  \textbf{TTRNN} \cite{yu2017long} and  Gated Recurrent Units (\textbf{GRU}) \cite{cho2014learning};   iv)  the two matrix/tensor-based  methods:  
\textbf{TRMF} \cite{yu2016temporal},   
 and 
\textbf{MOAR};   In addition, we combine MDT with MOAR by using  our obtained block Hankel tensor as the input of   MOAR   to get 
\textbf{BHT+MOAR} to evaluate the effectiveness of MDT together with tensor decomposition.

\subsubsection{Parameter settings}
All datasets are split into training sets (90$\%$) and testing sets (10$\%$).   We conduct grid search over parameters for each model and dataset (See Appendix B.1 of Supp. in detail). For our BHT-ARIMA, we will show its analysis of parameters in the following.  
We measure the forecasting accuracy  using the widely used Normalized Root Mean Square Error (\textbf{NRMSE})    metric.

\subsection {Analysis of  Parameters  and	Convergence }
We here not only  analyze the parameter sensitivity of   BHT-ARIMA,  but also    study the effects of   BHT-ARIMA  with  \textit{relaxed-orthogonality} and \textit{MDT on all  modes} by testing on the {Raw materials} and Smoke video datasets, respectively.

\subsubsection {Sensitivity Analysis of Parameters $\tau$, $\{R_m\}$ and  $(p,d,q)$}
Fig. $3$, $4$ and $5$ in the  Supp. show the forecasting results  using  BHT-ARIMA with  {full-orthogonality (\textbf{FO}) }  versus (\textbf{vs.})  {relaxed-orthogonality (\textbf{RO})}  with   different values of the  parameters  $\tau$, $\{R_m\}$ and  $(p,d,q)$, respectively.  Overall,   BHT-ARIMA with  RO is less sensitive and can achieve even slightly better forecasting accuracy than that  with FO.  Particularly, with   very small  $\tau$  (e.g. 1,2,5) and  when  the last mode rank $R_M =\tau $ (that maximally preserves the temporal dependencies among  core tensors),  BHT-ARIMA with  FO can achieve better  (similar) accuracy than that with RO.  With the same differencing  order $d$, the performance of BHT-ARIMA is relatively sensitive to too large $p$. Besides, the time cost of  BHT-ARIMA with RO is larger than   that of with FO  due to the cost of computing Eq.  \eqref{HTARIMA_UpdateUM} is larger than that of Eq.  \eqref{HTARIMA_UpdateU}.  

In short,    we do not need to carefully tune the parameters for BHT-ARIMA  while we usually  can obtain better results by  setting  smaller values such as  \{ $\tau = {2-4}$,   $R_M =\tau $ and small  $(p={1-5}, d={1},q={1})$\}.   Moreover,  the Tucker-rank can be estimated automatically \cite{yokota2016robust,shi2017tensor}.            

\subsubsection {Effect of  applying MDT on all modes /  temporal mode}
As discussed in \textbf{Remark 1} about why we apply MDT only along the temporal mode,  we here  verify it by testing on the Smoke video. As shown in  \textit{Fig. $6$ in the Supp.}: both types of applying MDT obtain similar forecasting accuracy while   using MDT on all the modes costs more time due to computing higher-order tensors.   This conclusion is also  applicable for the cases of  BHT-ARIMA with RO. These results support our assumption that  these TS items usually do not have strong neighborhood relationships   so it is unnecessary to applying MDT on other modes besides the temporal  mode.  
\begin{figure} [ttt!]
	\centering
\includegraphics[width=0.85\columnwidth]{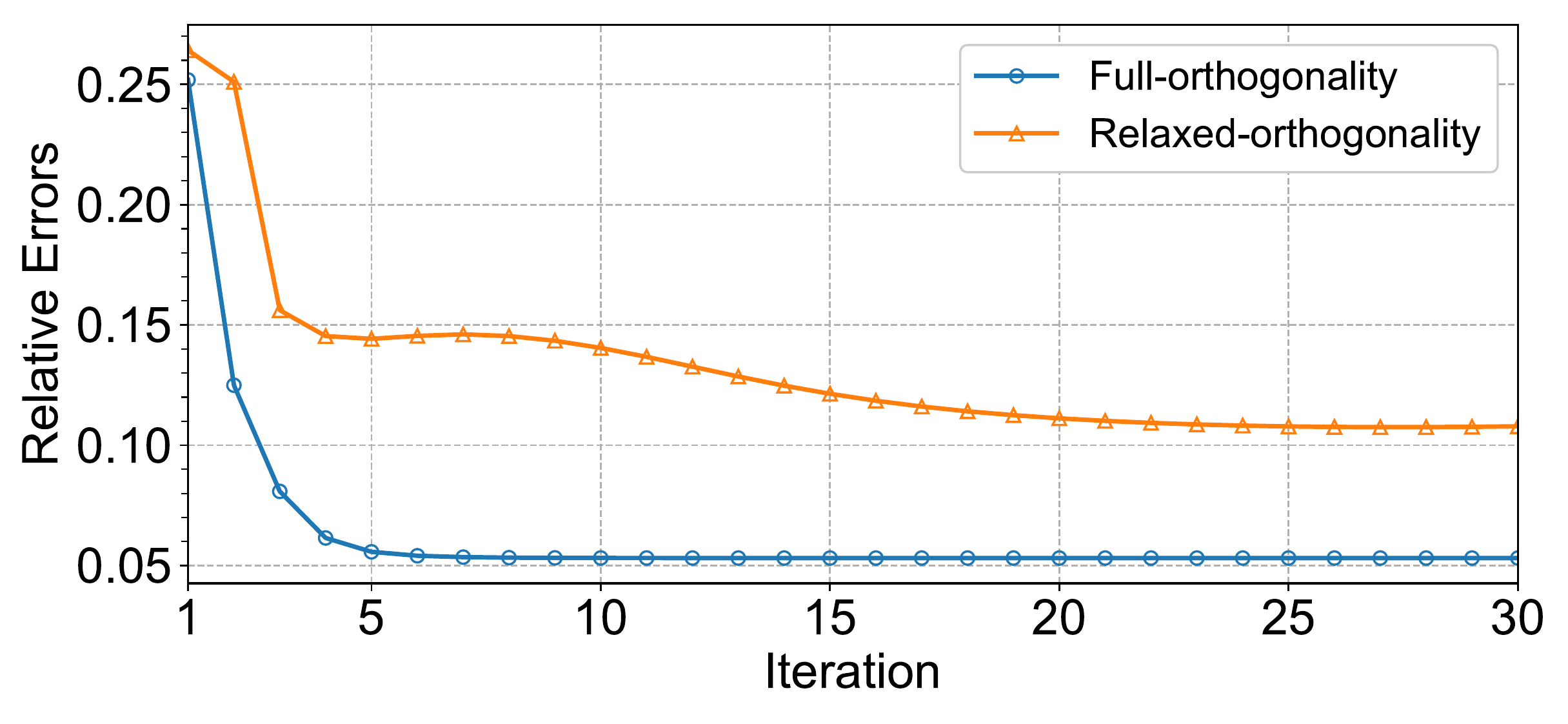} 	
\caption{\label{Convergence}Convergence curves  of BHT-ARIMA on   the Raw materials dataset.}
\end{figure}

\subsubsection {Convergence and  maximum iteration $K$}
We study the convergence of BHT-ARIMA
in terms of the relative error  of projection matrices $ \scriptsize \frac{ \sum_{m=1}^{M} ||\widehat{ \mathbf U}{^{(m)}}^{k+1}  -\widehat{ \mathbf U}{^{(m)}}^{k} ||_F^2} {\sum_{m=1}^{M}||\widehat{ \mathbf U}{^{(m)}}^{k+1}||_F^2 }$. 
Fig. \ref{Convergence}  shows that both  BHT-ARIMA     with  FO  and RO converge  quickly  while  the FO version  converges more smoothly and faster within 10 iterations. 
Furthermore, setting maximum iteration  $K>5$ is enough to get a sufficient forecasting accuracy, as shown   in \textit{Fig. $7(b)$ in the Supp..}
In this paper, we set $K=10$  for  BHT-ARIMA for all the tests.

\subsection{Forecasting Accuracy Comparison} 

We report the  average forecasting results of 10 runs in Tables \ref{TrafElec}, \ref{TablePCRM} and  \ref{TableTSFtimecost},  where we highlight \textbf{the best results} in bold font and underline \underline{the second best results}. Note that BHT-ARIMA used in the following comparisons is the full-orthogonality version. 

\begin{table}[ttt!] 
	\centering
	\setlength{\tabcolsep}{6.0pt}
	\renewcommand{\arraystretch}{1.3}
	\caption{ Forecasting results  comparison  on the Electricity and Traffic (measured in NRMSE).}
	
	\begin{mytabular2}{l|cccc}
		\hline
		& \begin{tabular}[c]{@{}c@{}}Electricity1096 \\ ( $\times  10^{-2} $)\end{tabular} & \begin{tabular}[c]{@{}c@{}}Electricity40\\ ($ \times 10^{-2} $)\end{tabular} & \begin{tabular}[c]{@{}c@{}}Traffic80\\ ($\times  10^{-3} $)\end{tabular} & \begin{tabular}[c]{@{}c@{}}Traffic40\\ ($ \times 10^{-3} $)\end{tabular} \\ \toprule 			
		ARIMA & \underline{1.316} & 10.091 & {3.194} & 6.097 \\\hline
		VAR  & 12.850 & {1.834} & {6.649}  & {1.526}\\ \hline
		XGBoost & 1.839 & 2.724 & 4.900 & 3.727 \\\hline
		Prophet & 13.799 & 5.948 & 4.343 & {1.992} \\\hline
		DeepAR & 4.742 & 4.857 & 8.178 & 5.358 \\\hline
		TTRNN & 2.686 & 3.565 & 5.723 & 3.432 \\\hline
		GRU  & 1.574  & \underline{1.545} & \underline{1.371}  & \underline{0.782}\\ \hline 
		TRMF & 6.221 & {2.713} & 5.800 & 2.340 \\\hline
		MOAR & 4.731 & 7.677 & 10.689 & 12.200 \\\hline
		BHT+MOAR & 3.787 & 4.050 & 4.920 & 4.464 \\ \hline \hline	
		\textbf{BHT-ARIMA} & \textbf{1.114} & \textbf{1.456} & \textbf{0.599} & \textbf{0.493}\\ \toprule
	\end{mytabular2} \label{TrafElec}  
\end{table}

\begin{table}[ttt!]
	\centering
	\setlength{\tabcolsep}{10.0pt}
	\renewcommand{\arraystretch}{1.3}
	\caption{Forecasting results   comparison on the PC sales and Raw materials (measured in NRMSE). }
	\begin{mytabular2}{l|ccc}
		\hline
		& \begin{tabular}[c]{@{}c@{}}PC sales \\($9 \hspace{-0.05cm}\times \hspace{-0.05cm}  105$) 
		\end{tabular} 
		& \begin{tabular}[c]{@{}c@{}}Raw materials$\_$I\\ ($1533 \hspace{-0.05cm}\times \hspace{-0.05cm}   24$)
		\end{tabular}& \begin{tabular}[c]{@{}c@{}}Raw materials$\_$II \\  ($2246 \hspace{-0.05cm}\times \hspace{-0.05cm}   24$)
		\end{tabular} 
		\\ \hline \hline
		ARIMA & 0.604 & 3.217 & 3.836  \\\hline
		VAR  & 0.690 & 3. 387 & 4.033  \\\hline 
		XGBoost & 0.618 & 3.834 & 4.231  \\\hline
		Prophet & {0.593}  & 2.984 & 3.734 \\\hline
		DeepAR & 0.689 & 3.158 & 4.476  \\\hline
		TTRNN  & 0.616  & {2.828} & {3.373}\\\hline		
		GRU  & \underline{0.524} & 2.592 & 3.250  \\\hline 
		TRMF  & 0.689  & 3.167 & 4.362\\\hline
		MOAR  & 0.689  &
		2.207 & 	2.635\\ \hline
		BHT+MOAR  & 0.683 & \underline{2.114} &	\underline{2.525}
		\\ \hline \hline
		BHT-ARIMA & \textbf{0.490} & \textbf{1.558} & \textbf{1.856} \\ \toprule
	\end{mytabular2} \label{TablePCRM}
\end{table}

\begin{table*}[htbp!]
	\centering
	\setlength{\tabcolsep}{7.8pt}
	\renewcommand{\arraystretch}{1.65} 
	\caption{Computational   cost  (seconds) vs. SOTA algorithms on all five real-world TS datasets. }
	\begin{mytabular2}{l|cccc|ccc|c}
		\hline
		Time(s) & \begin{tabular}[c]{@{}c@{}}
			Electricity1096\\ $(321 \hspace{-0.05cm}\times \hspace{-0.05cm}  1096) $\end{tabular} & \begin{tabular}[c]{@{}c@{}}Electricity40  \\  $(321 \hspace{-0.05cm}\times \hspace{-0.05cm}   40) $  \end{tabular}  & \begin{tabular}[c]{@{}c@{}}Traffic80  \\ $(228 \hspace{-0.05cm}\times \hspace{-0.05cm} 80 )$ \end{tabular}  
		& \begin{tabular}[c]{@{}c@{}}Traffic40    \\ $(228 \hspace{-0.08cm}\times \hspace{-0.08cm} 40 )$ \end{tabular}   
		& \begin{tabular}[c]{@{}c@{}}  PC sales \\    $(9 \hspace{-0.05cm}\times \hspace{-0.05cm} 105 )$ \end{tabular}& \begin{tabular}[c]{@{}c@{}}Raw materials$\_$I \\   $(1533 \hspace{-0.08cm}\times \hspace{-0.05cm}  24 )$ \end{tabular} & \begin{tabular}[c]{@{}c@{}}Raw  materials$\_$II \\ $(2246 \hspace{-0.05cm}\times \hspace{-0.05cm}  24 )$  \end{tabular} &\begin{tabular}[c]{@{}c@{}} Smoke Video \\ $(32 \hspace{-0.05cm}\times \hspace{-0.05cm}  64  \hspace{-0.05cm}\times \hspace{-0.05cm}  100)$  \end{tabular}  \\ \hline \hline
		ARIMA & 3322.62 & 327.60 & 453.63 & 382.32 & 13.01 &  1151.35 & 2067.48 & 1340.20 \\ \hline 
	VAR & 68.96 & 2.042 & 4.19 & \underline{0.92} & \textbf{0.27} &  11.32 & 25.51 & 69.63 \\ \hline 				
		XGBoost & 108.02 & 26.07 & 23.28 & 6.94 & 8.87 &  148.22 & 159.33 &  178.86 \\  \hline 
		Prophet & 2160.33 & 1304.07 & 1241.43 & 335.23 & 53.31 &  434.15 & 853.23 & 13813.21 \\ \hline 
		DeepAR & 175.56 & 171.17 & 119.02 & 112.04 &  78.43 & 136.80 & 286.66 &  106.65 \\  \hline 
			TTRNN & 165.84 & 104.38 & 88.77 & 95.47 &  22.83 & 25.49 & 28.05 & 27.82 \\ \hline
		GRU &  4383.81 & 1077.09 & 1104.17 &  858.03 & 93.34 & 3353.43  & 3534.96   & 3142.46 \\ \hline 				
		TRMF & \textbf{16.03} & {\textbf{0.57}} & \underline{1.90} & {1.22} &  {1.46} & \textbf{0.27}& \textbf{0.28} & \textbf{0.77} \\ \hline
		MOAR & 28.96 & 10.21 & 6.75 & 5.78  & 1.68 &  566.19 & 1612.52  &7.26 \\ \hline 
		BHT+MOAR & 42.86 & 12.28 & 8.28 & 5.93 
		&1.70 
		&  865.91 & 
		1700.49	& 10.95 \\ \hline \hline
		\textbf{BHT-ARIMA }& \underline{23.24} & \underline{0.89} & \textbf{1.28} & \textbf{0.58} & \underline{1.41} &
		\underline{0.91} & \underline{1.13}
		&  \underline{3.67} \\ \toprule
	\end{mytabular2} \label{TableTSFtimecost} 
\end{table*}

\subsubsection{Forecasting results of  longer  vs. shorter  TS}

To evaluate the capability of  BHT-ARIMA for longer vs. shorter TS, we  sample the first 40 time points of the Traffic dataset and thus get shorter \textbf{Traffic40}   ($228 \times 40$) and denote original whole set as    \textbf{Traffic80}. For the Electricity, we sample the  first 40 time points   and get   \textbf{Electricity40}   ($321 \times  40$), and the whole one is denoted as \textbf{Electricity1096}.   
The  results are reported in  \textbf{Table} \ref{TrafElec}:   BHT-ARIMA outperforms all the existing competing methods in all the cases.  Especially for shorter TS,  BHT-ARIMA shows more  advantage with  54.5$\%$ improvement on average on the Traffic data than that of on the Electricity (7.9$\%$ improvement on average). GRU and  ARIMA  share   the second best results.  Note that BHT+MOAR performs consistently better than MOAR in all cases, which verifies the effectiveness of  applying MDT together with tensor decomposition. 

\subsubsection{Forecasting results of  industrial TS datasets}
In practice, industrial datasets are more complex with  random irregular patterns and larger variance compared to these  public  datasets.  As reported in the \textbf{Table} \ref{TablePCRM}: 1) For PC sales with only nine TS items,  BHT-ARIMA   outperforms the second best performing method GRU  by  6.5$\%$ on average; 2) For Raw materials dataset which has shorter length (24 months) than PC sales (105 weeks) while has much larger number of  items (2246), we remove the items with missing values and get  1553 items namely \textbf{Raw materials$\_$I} while the original one named as \textbf{Raw materials$\_$II}.  In such scenarios,  BHT + MOAR and GRU perform better than other existing methods, although their  results are much worse than our BHT-ARIMA.   These results further confirm the effectiveness of MDT together with tensor decomposition and also verified the improvement of explicitly using the core tensors to train our ARIMA model.

\subsubsection{Forecasting results of Smoke video} 
\begin{figure}[ttt!]    
	\centering
	\includegraphics[width=0.8\columnwidth ]{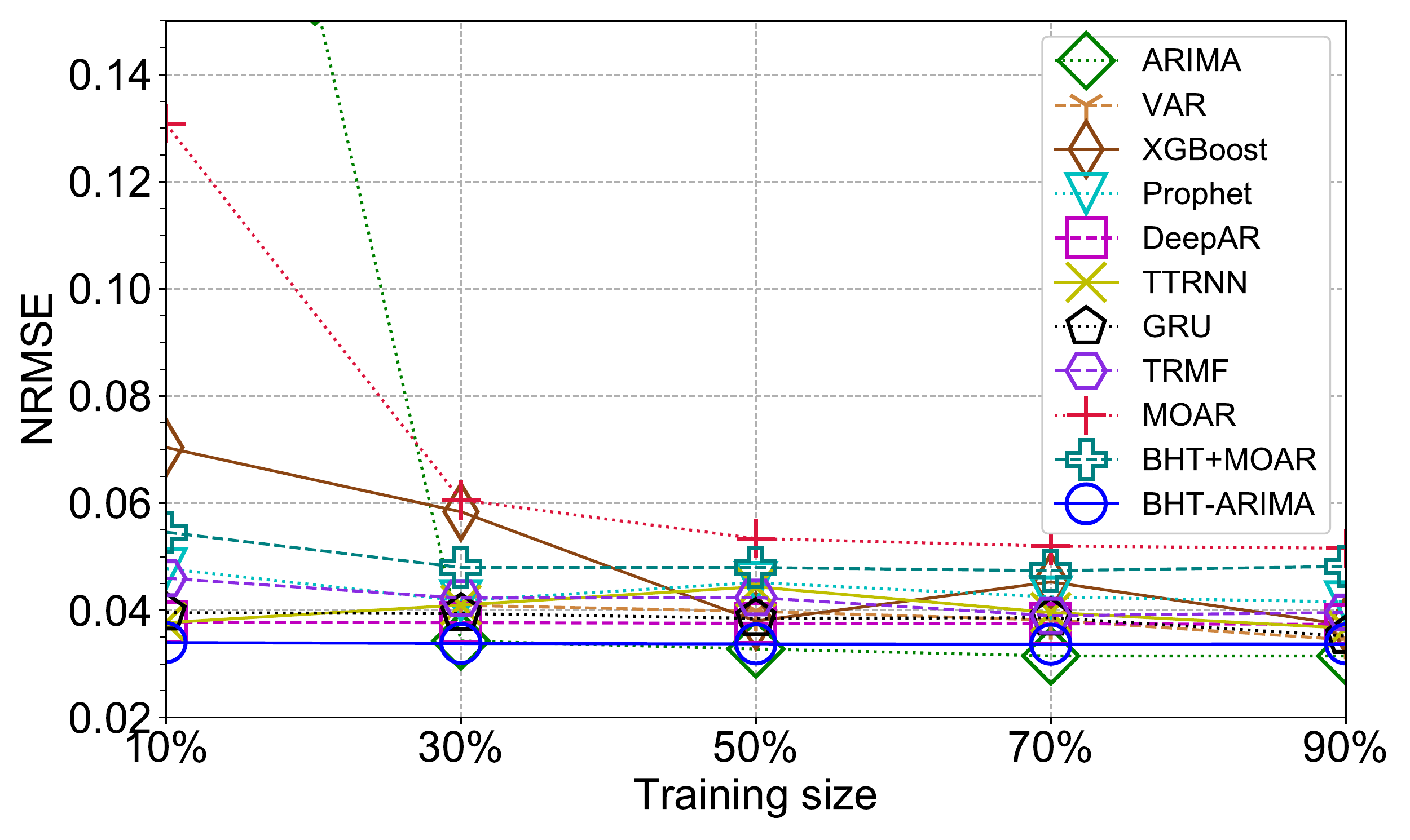}
	\caption{\label{Figvideo}Forecasting accuracy comparison on the Smoke video. In the case of using $10\%$ training size,   the  NRMSE of ARIMA is too large  (0.277) so we truncate its curve there.}
\end{figure}

We further evaluate the performance of BHT-ARIMA on higher-order TS data using the  Smoke video with $10\% - 90\%$ training set.  Fig. \ref{Figvideo} shows that BHT-ARIMA consistently forecasts the next video frame with  smaller errors using even $10\%$ training data (10 frames) where ARIMA, MOAR and other methods fail to keep their  performance. Moreover,  although ARIMA can achieve  slightly better accuracy than ours  with  more than $50\%$ training data,  it surfers from extremely larger computational cost and memory requirements because like other  linear models who cannot directly handle  tensor TS data  and  need to reshape  them into vectors. 

\begin{figure}[h]
	\centering
	\subfloat[Traffic80]{\includegraphics[ width=0.505\columnwidth]{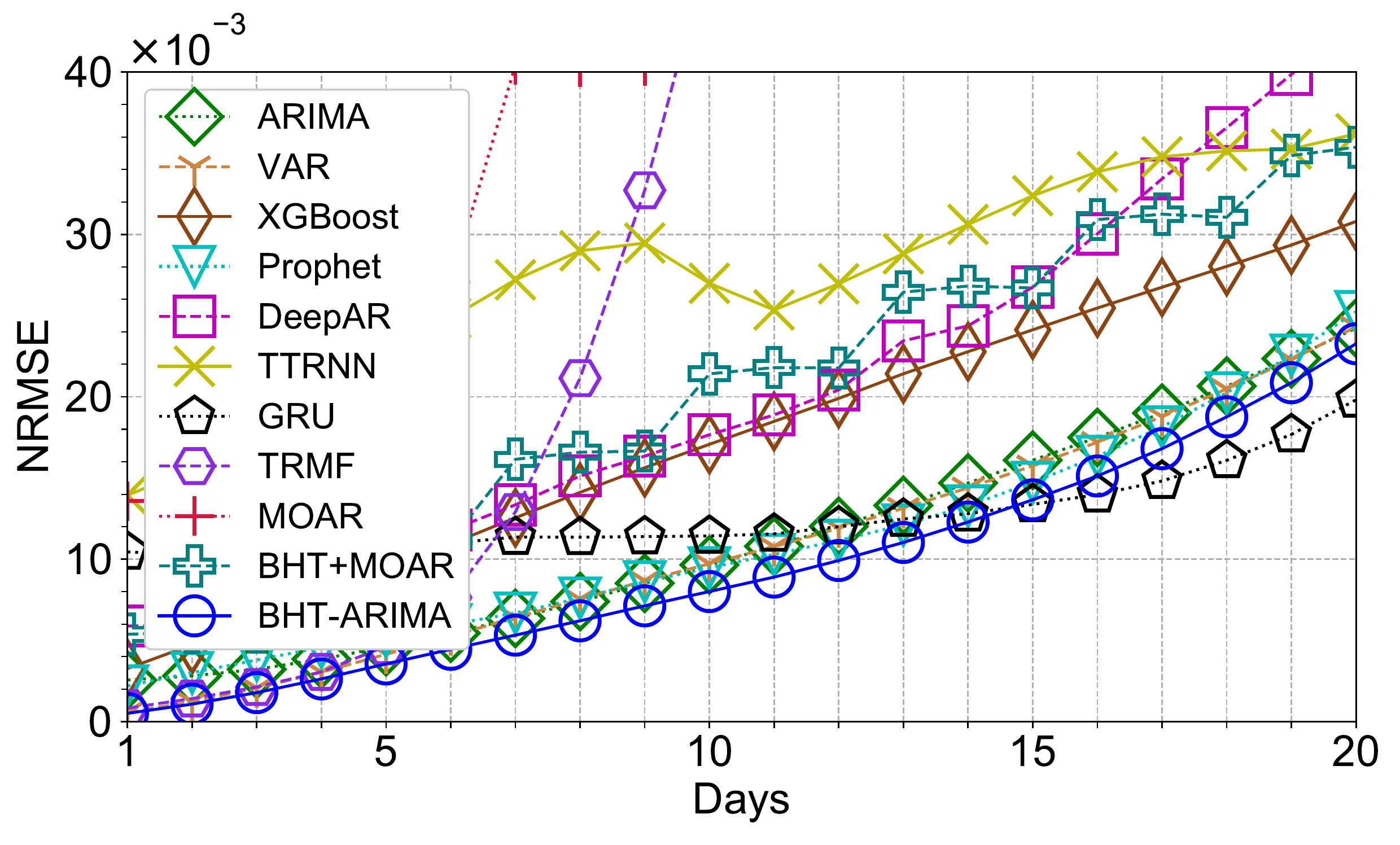}} 
	\label{Multistep_Traffic80} 
	\subfloat[Raw  materials$\_$II]{\includegraphics[ width=0.485\columnwidth]{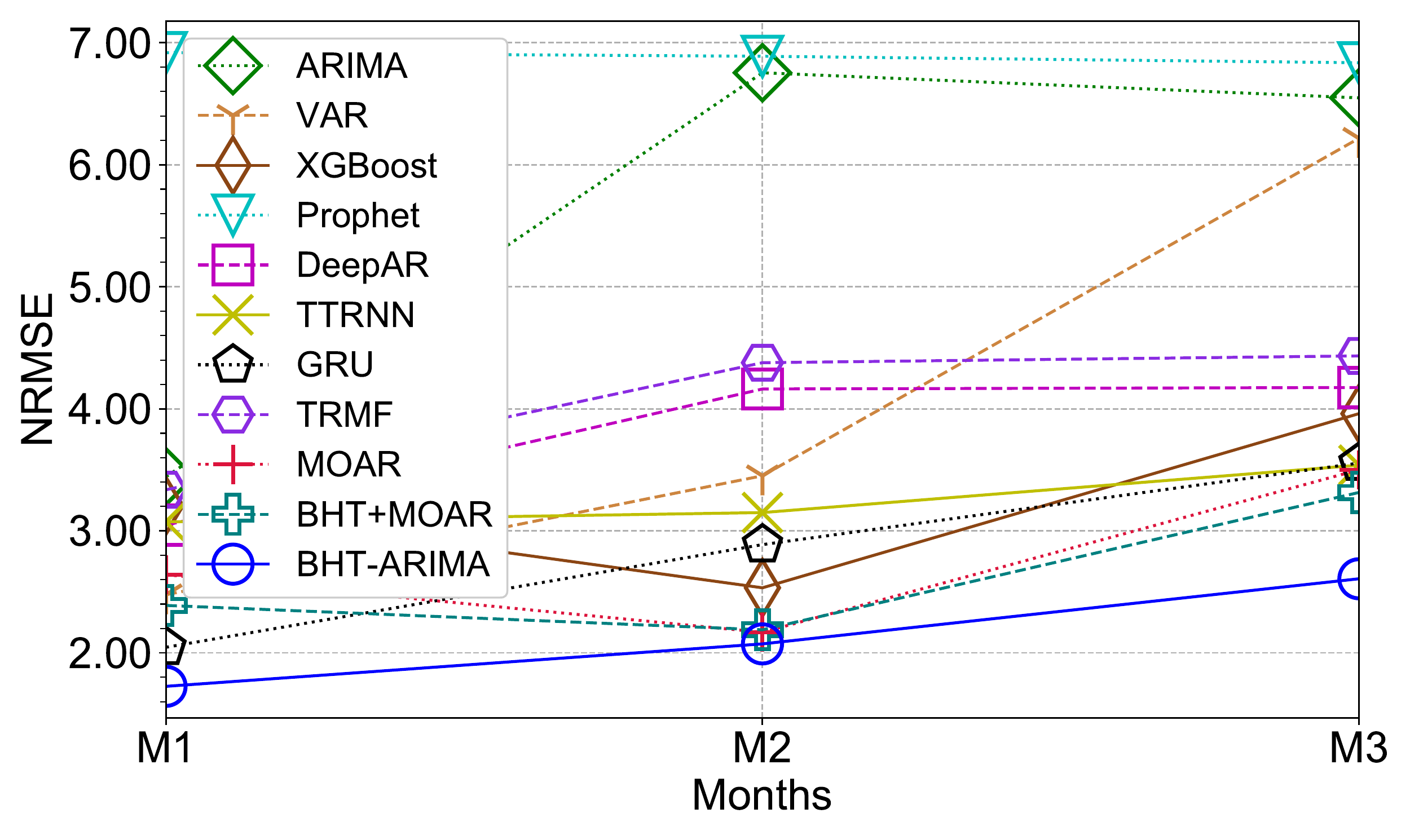}} 
	 \label{Multistep_Raw}
	\caption{\label{raw2246_multistep} Long-term forecasting  results for the Traffic  and Raw materials  (truncate curves with too large NRMSE).}
\end{figure}

\subsection{Long-term Forecasting  Comparison} 

Fig. \ref{raw2246_multistep}  shows the comparison of long-term forecasting results, which further confirm the promising performance of BHT-ARIMA. Forecasting  more  steps, the errors of all the  methods  increase in general while BHT-ARIMA consistently keeps its best performance on the whole, especially on the shorter Raw materials  dataset. Note that GRU slightly outperforms our method after  15 steps on Traffic80 dataset, but it requires near 900  times computational cost than ours.     

\subsection{Time Cost Comparison} 
We report the average time cost of forecasting in \textbf{Table}  \ref{TableTSFtimecost}.    Although TRMF is slightly slower   than VAR in a few cases,  it is the fastest algorithm on the whole due to its core parts are implemented by C programming. BHT-ARIMA  is the second fastest method while our implementations are not optimized for efficiency as our focus here is accuracy. MOAR is slower than ours mainly because it does not directly use the low-dimensional core tensors for training.  ARIMA and GRU are  the second best performing  algorithms in a few cases, but  they are the most slowest methods (more than  500 times slower than BHT-ARIMA on average).

\section{Conclusions}

In this paper, we  proposed a novel  BHT-ARIMA for multiple (short) TSF. BHT-ARIMA tactically utilizes the unique strengths of  smart tensorization via MDT, tensor ARIMA, and low-rank Tucker 
decomposition in a unified model.  With low-rank Hankel tensor in embedded space by MDT Hankelization along  the temporal mode, we further obtain the compressed core tensors using Tucker decomposition. The core tensors   capture the intrinsic correlations among multiple TS and  are explicitly used to train the  tensor ARIMA model.  BHT-ARIMA   can improve  forecasting accuracy and computational speed, especially for multiple short TS.  We also empirically studied its robustness  to various parameters by comparing it with its relaxed-orthogonality version.  Experiments conducted on five real-world TS datasets demonstrate that  BHT-ARIMA outperforms  the SOTA methods with significant improvement.  

\section*{Acknowledgments}
This  research was partially supported by the Ministry of Education and Science of the Russian Federation (grant 14.756.31.0001) and JST ACT-I: Grant Number JPMJPR18UU. The authors  would like to thank Dr. Peiguang Jing for his helpful discussions. 

\bibliography{8407_bib}

\begin{thebibliography}{}

\bibitem[\protect\citeauthoryear{Agarwal \bgroup et al\mbox.\egroup
  }{2018}]{agarwal2018model}
Agarwal, A.; Amjad, M.~J.; Shah, D.; and Shen, D.
\newblock 2018.
\newblock Model agnostic time series analysis via matrix estimation.
\newblock {\em POMACS} 2(3):40.

\bibitem[\protect\citeauthoryear{Bhanu \bgroup et al\mbox.\egroup
  }{2018}]{bhanu2018forecasting}
Bhanu, M.; Priya, S.; Dandapat, S.~K.; Chandra, J.; and Mendes-Moreira, J.
\newblock 2018.
\newblock Forecasting traffic flow in big cities using modified {Tucker}
  decomposition.
\newblock In {\em ADMA},  119--128.
\newblock Springer.

\bibitem[\protect\citeauthoryear{Box and Jenkins}{1968}]{box1968some}
Box, G.~E., and Jenkins, G.~M.
\newblock 1968.
\newblock Some recent advances in forecasting and control.
\newblock {\em Journal of the Royal Statistical Society. Series C (Applied
  Statistics)} 17(2):91--109.

\bibitem[\protect\citeauthoryear{Chen and Guestrin}{2016}]{chen2016xgboost}
Chen, T., and Guestrin, C.
\newblock 2016.
\newblock Xgboost: A scalable tree boosting system.
\newblock In {\em ACM SIGKDD},  785--794.
\newblock ACM.

\bibitem[\protect\citeauthoryear{Chen \bgroup et al\mbox.\egroup
  }{2018}]{chen2018neucast}
Chen, P.; Liu, S.; Shi, C.; Hooi, B.; Wang, B.; and Cheng, X.
\newblock 2018.
\newblock {NeuCast}: seasonal neural forecast of power grid time series.
\newblock In {\em IJCAI},  3315--3321.
\newblock AAAI Press.

\bibitem[\protect\citeauthoryear{Cho \bgroup et al\mbox.\egroup
  }{2014}]{cho2014learning}
Cho, K.; Van~Merri{\"e}nboer, B.; Gulcehre, C.; Bahdanau, D.; Bougares, F.;
  Schwenk, H.; and Bengio, Y.
\newblock 2014.
\newblock Learning phrase representations using rnn encoder-decoder for
  statistical machine translation.
\newblock {\em arXiv preprint arXiv:1406.1078}.

\bibitem[\protect\citeauthoryear{Cichocki \bgroup et al\mbox.\egroup
  }{2016}]{cichocki2016tensor}
Cichocki, A.; Lee, N.; Oseledets, I.; Phan, A.-H.; Zhao, Q.; Mandic, D.~P.;
  et~al.
\newblock 2016.
\newblock Tensor networks for dimensionality reduction and large-scale
  optimization: Part 1 low-rank tensor decompositions.
\newblock {\em Found. Trends{\textregistered} Mach. Learn.} 9(4-5):249--429.

\bibitem[\protect\citeauthoryear{de Araujo, Ribeiro, and
  Faloutsos}{2017}]{de2017tensorcast}
de~Araujo, M.~R.; Ribeiro, P. M.~P.; and Faloutsos, C.
\newblock 2017.
\newblock {Tensorcast}: Forecasting with context using coupled tensors.
\newblock In {\em ICDM},  71--80.
\newblock IEEE.

\bibitem[\protect\citeauthoryear{Ding, Qi, and Wei}{2015}]{ding2015fast}
Ding, W.; Qi, L.; and Wei, Y.
\newblock 2015.
\newblock Fast {Hankel} tensor--vector product and its application to
  exponential data fitting.
\newblock {\em Numer. Linear Algebra Appl.} 22(5):814--832.

\bibitem[\protect\citeauthoryear{Dunlavy, Kolda, and
  Acar}{2011}]{dunlavy2011temporal}
Dunlavy, D.~M.; Kolda, T.~G.; and Acar, E.
\newblock 2011.
\newblock Temporal link prediction using matrix and tensor factorizations.
\newblock {\em ACM Trans. Knowl. Discov. Data} 5(2):10.

\bibitem[\protect\citeauthoryear{Faloutsos \bgroup et al\mbox.\egroup
  }{2018}]{faloutsos2018forecasting}
Faloutsos, C.; Gasthaus, J.; Januschowski, T.; and Wang, Y.
\newblock 2018.
\newblock Forecasting big time series: old and new.
\newblock {\em Proceedings of the VLDB Endowment} 11(12):2102--2105.

\bibitem[\protect\citeauthoryear{Faloutsos \bgroup et al\mbox.\egroup
  }{2019}]{faloutsos2019forecasting}
Faloutsos, C.; Flunkert, V.; Gasthaus, J.; Januschowski, T.; and Wang, Y.
\newblock 2019.
\newblock Forecasting big time series: Theory and practice.
\newblock In {\em ACM SIGKDD},  3209--3210.
\newblock ACM.

\bibitem[\protect\citeauthoryear{Fanaee-T and Gama}{2016}]{fanaee2016tensor}
Fanaee-T, H., and Gama, J.
\newblock 2016.
\newblock Tensor-based anomaly detection: An interdisciplinary survey.
\newblock {\em Knowledge-Based Systems} 98:130--147.

\bibitem[\protect\citeauthoryear{Higham and
  Papadimitriou}{1995}]{higham1995matrix}
Higham, N., and Papadimitriou, P.
\newblock 1995.
\newblock Matrix {P}rocrustes {P}roblems.
\newblock {\em Rapport technique, University of Manchester}.

\bibitem[\protect\citeauthoryear{Jing \bgroup et al\mbox.\egroup
  }{2018}]{jing2018high}
Jing, P.; Su, Y.; Jin, X.; and Zhang, C.
\newblock 2018.
\newblock High-order temporal correlation model learning for time-series
  prediction.
\newblock {\em IEEE Trans. Cybern.} 49(6):2385--2397.

\bibitem[\protect\citeauthoryear{Khashei and Bijari}{2011}]{khashei2011novel}
Khashei, M., and Bijari, M.
\newblock 2011.
\newblock A novel hybridization of artificial neural networks and {ARIMA}
  models for time series forecasting.
\newblock {\em Applied Soft Computing} 11(2):2664--2675.

\bibitem[\protect\citeauthoryear{Kolda and Bader}{2009}]{kolda2009tensor}
Kolda, T.~G., and Bader, B.~W.
\newblock 2009.
\newblock Tensor decompositions and applications.
\newblock {\em SIAM Rev.} 51(3):455--500.

\bibitem[\protect\citeauthoryear{Lai \bgroup et al\mbox.\egroup
  }{2018}]{lai2018modeling}
Lai, G.; Chang, W.-C.; Yang, Y.; and Liu, H.
\newblock 2018.
\newblock Modeling long-and short-term temporal patterns with deep neural
  networks.
\newblock In {\em ACM SIGIR},  95--104.
\newblock ACM.

\bibitem[\protect\citeauthoryear{Li \bgroup et al\mbox.\egroup
  }{2015}]{li2015tensor}
Li, Q.; Jiang, L.; Li, P.; and Chen, H.
\newblock 2015.
\newblock Tensor-based learning for predicting stock movements.
\newblock In {\em AAAI},  1784--1790.
\newblock AAAI Press.

\bibitem[\protect\citeauthoryear{Liu \bgroup et al\mbox.\egroup
  }{2016}]{liu2016online}
Liu, C.; Hoi, S.~C.; Zhao, P.; and Sun, J.
\newblock 2016.
\newblock Online {ARIMA} algorithms for time series prediction.
\newblock In {\em AAAI},  1867--1873.
\newblock AAAI Press.

\bibitem[\protect\citeauthoryear{Ma \bgroup et al\mbox.\egroup
  }{2019}]{ma2019large}
Ma, X.; Zhang, L.; Xu, L.; Liu, Z.; Chen, G.; Xiao, Z.; Wang, Y.; and Wu, Z.
\newblock 2019.
\newblock Large-scale user visits understanding and forecasting with deep
  spatial-temporal tensor factorization framework.
\newblock In {\em ACM SIGKDD},  2403--2411.
\newblock ACM.

\bibitem[\protect\citeauthoryear{Rogers, Li, and
  Russell}{2013}]{rogers2013multilinear}
Rogers, M.; Li, L.; and Russell, S.~J.
\newblock 2013.
\newblock Multilinear dynamical systems for tensor time series.
\newblock In {\em NeurIPS},  2634--2642.

\bibitem[\protect\citeauthoryear{Salinas, Flunkert, and
  Gasthaus}{2017}]{salinas2017deepar}
Salinas, D.; Flunkert, V.; and Gasthaus, J.
\newblock 2017.
\newblock {DeepAR}: Probabilistic forecasting with autoregressive recurrent
  networks.
\newblock {\em arXiv preprint arXiv:1704.04110}.

\bibitem[\protect\citeauthoryear{Shi \bgroup et al\mbox.\egroup
  }{2018}]{shi2018feature}
Shi, Q.; Cheung, Y.-M.; Zhao, Q.; and Lu, H.
\newblock 2018.
\newblock Feature extraction for incomplete data via low-rank tensor
  decomposition with feature regularization.
\newblock {\em IEEE Trans. Neural Netw. Learn. Syst.} 30(6):1803--1817.

\bibitem[\protect\citeauthoryear{Shi, Lu, and Cheung}{2017}]{shi2017tensor}
Shi, Q.; Lu, H.; and Cheung, Y.-m.
\newblock 2017.
\newblock Tensor rank estimation and completion via {CP}-based nuclear norm.
\newblock In {\em CIKM},  949--958.
\newblock ACM.

\bibitem[\protect\citeauthoryear{Smyl and Kuber}{2016}]{smyl2016data}
Smyl, S., and Kuber, K.
\newblock 2016.
\newblock Data preprocessing and augmentation for multiple short time series
  forecasting with recurrent neural networks.
\newblock In {\em ISF}.
\newblock Santander.

\bibitem[\protect\citeauthoryear{Sun and Chen}{2019}]{sun2019bayesian}
Sun, L., and Chen, X.
\newblock 2019.
\newblock Bayesian temporal factorization for multidimensional time series
  prediction.
\newblock {\em arXiv preprint arXiv:1910.06366}.

\bibitem[\protect\citeauthoryear{Tan \bgroup et al\mbox.\egroup
  }{2016}]{tan2016short}
Tan, H.; Wu, Y.; Shen, B.; Jin, P.~J.; and Ran, B.
\newblock 2016.
\newblock Short-term traffic prediction based on dynamic tensor completion.
\newblock {\em IEEE Trans. Intell. Transp. Syst.} 17(8):2123--2133.

\bibitem[\protect\citeauthoryear{Taylor and
  Letham}{2018}]{taylor2018forecasting}
Taylor, S.~J., and Letham, B.
\newblock 2018.
\newblock Forecasting at scale.
\newblock {\em The American Statistician} 72(1):37--45.

\bibitem[\protect\citeauthoryear{Yokota and Hontani}{2018}]{yokota2018tensor}
Yokota, T., and Hontani, H.
\newblock 2018.
\newblock Tensor completion with shift-invariant cosine bases.
\newblock In {\em APSIPA ASC},  1325--1333.
\newblock IEEE.

\bibitem[\protect\citeauthoryear{Yokota \bgroup et al\mbox.\egroup
  }{2018}]{yokota2018missing}
Yokota, T.; Erem, B.; Guler, S.; Warfield, S.~K.; and Hontani, H.
\newblock 2018.
\newblock Missing slice recovery for tensors using a low-rank model in embedded
  space.
\newblock In {\em CVPR},  8251--8259.

\bibitem[\protect\citeauthoryear{Yokota \bgroup et al\mbox.\egroup
  }{2019}]{yokota2019manifold}
Yokota, T.; Hontani, H.; Zhao, Q.; and Cichocki, A.
\newblock 2019.
\newblock Manifold modeling in embedded space: A perspective for interpreting
  ``deep image prior".
\newblock {\em arXiv preprint arXiv:1908.02995}.

\bibitem[\protect\citeauthoryear{Yokota, Lee, and
  Cichocki}{2016}]{yokota2016robust}
Yokota, T.; Lee, N.; and Cichocki, A.
\newblock 2016.
\newblock Robust multilinear tensor rank estimation using higher order singular
  value decomposition and information criteria.
\newblock {\em IEEE Trans. Signal Process.} 65(5):1196--1206.

\bibitem[\protect\citeauthoryear{Yu \bgroup et al\mbox.\egroup
  }{2017}]{yu2017long}
Yu, R.; Zheng, S.; Anandkumar, A.; and Yue, Y.
\newblock 2017.
\newblock Long-term forecasting using {T}ensor-{T}rain {RNN}s.
\newblock {\em arXiv preprint arXiv:1711.00073}.

\bibitem[\protect\citeauthoryear{Yu, Rao, and Dhillon}{2016}]{yu2016temporal}
Yu, H.-F.; Rao, N.; and Dhillon, I.~S.
\newblock 2016.
\newblock Temporal regularized matrix factorization for high-dimensional time
  series prediction.
\newblock In {\em NeurIPS},  847--855.

\bibitem[\protect\citeauthoryear{Yu, Yin, and Zhu}{2017}]{yu2017spatio}
Yu, B.; Yin, H.; and Zhu, Z.
\newblock 2017.
\newblock Spatio-temporal graph convolutional networks: A deep learning
  framework for traffic forecasting.
\newblock {\em arXiv preprint arXiv:1709.04875}.

\bibitem[\protect\citeauthoryear{Zhang}{2003}]{zhang2003time}
Zhang, G.~P.
\newblock 2003.
\newblock Time series forecasting using a hybrid {ARIMA} and neural network
  model.
\newblock {\em Neurocomputing} 50:159--175.

\bibitem[\protect\citeauthoryear{Zhou and Cheung}{2019}]{zhou2019bayesian}
Zhou, Y., and Cheung, Y.
\newblock 2019.
\newblock Bayesian low-tubal-rank robust tensor factorization with multi-rank
  determination.
\newblock {\em IEEE Trans. Pattern Anal. Mach. Intell.} (In Press).

\bibitem[\protect\citeauthoryear{Zhou, Lu, and
  Cheung}{2019}]{zhou2019probabilistic}
Zhou, Y.; Lu, H.; and Cheung, Y.-M.
\newblock 2019.
\newblock Probabilistic rank-one tensor analysis with concurrent
  regularizations.
\newblock {\em IEEE Trans. Cybern.} (In Press).

\end{thebibliography}
\bibliographystyle{aaai}
\end{document}